\documentclass{article}

\usepackage{arxiv}

\usepackage[utf8]{inputenc} 
\usepackage[T1]{fontenc}    
\usepackage{hyperref}       
\usepackage{url}            
\usepackage{booktabs}       
\usepackage{amsfonts}       
\usepackage{nicefrac}       
\usepackage{microtype}      
\usepackage{lipsum}		
\usepackage{graphicx}
\usepackage[numbers]{natbib}
\usepackage{doi}

\usepackage{comment}
\usepackage{color, colortbl}
\usepackage{float}
\usepackage{soul}
\usepackage{arydshln} 
\usepackage{booktabs}
\usepackage{multirow}
\usepackage{array}
\usepackage{geometry}
\usepackage{enumitem}

\usepackage{xcolor}

\definecolor{Gray}{gray}{0.9}
\definecolor{LightCyan}{rgb}{0.88,1,1}

\usepackage{mdframed} 

\usepackage{framed}
\newenvironment{prompt}
  {\begin{leftbar}\small}
  {\end{leftbar}}

\title{Can GPT models Follow Human Summarization Guidelines? A Study for Targeted Communication Goals}

\date{} 					

\author{ 
	{Yongxin ZHOU, Fabien RINGEVAL, and François PORTET} \\
Univ. Grenoble Alpes, CNRS, Inria, Grenoble INP, LIG, 38000 Grenoble, France \\
firstname.lastname@univ-grenoble-alpes.fr \\
}

\renewcommand{\shorttitle}{Can GPT models Follow Human Summarization Guidelines?}

\hypersetup{
pdftitle={Can GPT models Follow Human Summarization Guidelines? A Study for Targeted Communication Goals},
pdfauthor={Yongxin ZHOU, Fabien RINGEVAL, and François PORTET},
pdfkeywords={Large Language Models (LLMs), Human Summarization Guidelines, Dialogue Summarization},
}

\begin{document}
\maketitle
\begin{abstract}
This study investigates the ability of GPT models (ChatGPT, GPT-4 and GPT-4o) to generate dialogue summaries that adhere to human guidelines. Our evaluation involved experimenting with various prompts to guide the models in complying with guidelines on two datasets: DialogSum (English social conversations) and DECODA (French call center interactions). 
Human evaluation, based on summarization guidelines, served as the primary assessment method, complemented by extensive quantitative and qualitative analyses. 
Our findings reveal a preference for GPT-generated summaries over those from task-specific pre-trained models and reference summaries, highlighting GPT models' ability to follow human guidelines despite occasionally producing longer outputs and exhibiting divergent lexical and structural alignment with references.
The discrepancy between ROUGE, BERTScore, and human evaluation underscores the need for more reliable automatic evaluation metrics.\footnote{The generated summaries and human annotations are publicly available at \url{https://github.com/yongxin2020/LLM-Sum-Guidelines}}
\end{abstract}


\newcommand{\chatgpt}{\texttt{ChatGPT}}
\newcommand{\gptfour}{\texttt{GPT-4}}
\newcommand{\gptfouro}{\texttt{GPT-4o}}

\section{Introduction}

Although instruction-tuned Large Language Models (LLMs) have shown impressive performance on several benchmark datasets, they still struggle with various challenging tasks \citep{laskar-etal-2023-systematic, qin-etal-2023-chatgpt}. On automatic metrics like ROUGE, general-purpose models (e.g., ChatGPT) consistently trail behind task-specific, fine-tuned models \citep{bang-etal-2023-multitask, zhang-etal-2023-extractive-summarization, yang2023exploring}. In dialogue summarization specifically, ChatGPT demonstrates poorer performance than state-of-the-art fine-tuned models on datasets including SAMSum \citep{gliwa-etal-2019-samsum} and DialogSum \citep{chen-etal-2021-dialogsum}.
However, human evaluation studies reveal a more nuanced picture: instruction-tuned LLMs can demonstrate human-aligned summarization capabilities that automatic metrics may fail to capture \citep{zhang-etal-2024-benchmarking}.

The underlying causes of GPT models' underperformance on automatic summarization metrics remain poorly understood.
Human summarization is a complex cognitive process that involves understanding, condensing, and conveying essential information while adhering to specific guidelines.
Given that human annotators have to follow these guidelines when creating reference summaries, we speculate that the performance discrepancy could be attributed to the lack of explicit summarization instructions tailored to specific communication goals during model prompting. This is particularly critical for dialogue summarization, where targeted communication goals play a significant role. 

Even though dialogue summarization is a well-established task, it remains challenging due to the absence of a consensus for what constitutes an ideal dialogue summary~\citep{guo-etal-2022-questioning}. The approach to dialogue summarization is heavily influenced by specific communication objectives, which vary across different contexts such as meetings, or customer service interactions.
A review of guidelines from major dialogue summarization datasets reveals that each corpus defines its own set of objectives for creating reference summaries, applying different summary criteria to meet specific needs~\citep{zhou-etal-2024-psentscore}.
For instance, in the context of customer service, while TWEETSUMM~\citep{feigenblat-etal-2021-tweetsumm-dialog} offers both extractive and abstractive summaries, CSDS~\citep{lin-etal-2021-csds} provides three distinct summaries for each dialogue: an overall summary and two role-oriented summaries (user and agent). 

In this work, we evaluated the ability of GPT models to generate dialogue summaries that adhere to human guidelines, which provides the following contributions: 

\begin{itemize}[noitemsep,topsep=0pt,parsep=0pt,partopsep=0pt]
    \item Developing a range of prompts, from simplified instructions 
    to detailed human-annotator guidelines, for summarization tasks; 
    \item Evaluating three GPT models against task-specific pre-trained models on English and French datasets with distinct communication goals (DialogSum and DECODA); 
    \item Assessing performance using automatic metrics (ROUGE, BERTScore, and LLM-as-judge) and a task-aligned human evaluation framework;
    \item Performing comprehensive analyses to assess and explain model performance, identifying both capabilities and limitations in meeting targeted communication objectives.
\end{itemize}

\section{Related Work}
\subsection{Evaluation of Summarization Tasks}
The dominant approach for automatic summarization evaluation employs similarity-based metrics. ROUGE \citep{lin-2004-rouge} remains the standard, measuring n-gram overlap (e.g., ROUGE-1, ROUGE-2, ROUGE-L) between system outputs and human references. 
To capture semantic similarity beyond lexical overlap, embedding-based metrics like BERTScore \citep{Zhang*2020BERTScore:} leverage contextual embeddings.
Alternative paradigms include natural language inference (NLI) for evaluating factual consistency through entailment \citep{10.1162/tacl_a_00576} and question answering (QA)-based methods that use question generation and answering \citep{wang-etal-2020-asking}.
More recently, LLM-based evaluation has emerged, encompassing metrics derived from, prompted by, or fine-tuned on LLMs \citep{10.1162/coli_a_00561}. For instance, G-Eval \citep{liu-etal-2023-g} operates without references and assesses dimensions like coherence, consistency, fluency, and relevance.

Beyond automatic metrics, human evaluation of summaries has evolved into a multi-dimensional framework assessing faithfulness, coherence, relevance, completeness, and conciseness \citep{fabbri-etal-2021-summeval}. 
This framework is applied across diverse domains: from general news benchmarks evaluating faithfulness, coherence, and relevance \citep{zhang-etal-2024-benchmarking}, to specialized tasks like topic-focused dialogue summarization assessed for completeness, relevance, and factual consistency \citep{tang-etal-2024-tofueval},
to domain-specific evaluations (e.g., bookings, interviews) that employ key fact validation to calculate scores for faithfulness, completeness, and conciseness \citep{min-etal-2025-towards}. 

Nevertheless, a number of challenges remain. Automatic metrics show limited correlation with human judgment \citep{fabbri-etal-2021-summeval}.
LLM-based approaches face challenges such as prompt sensitivity, limited confidence calibration, the need for human oversight (restricting full automation), and limited explainability \citep{10.1162/coli_a_00561}.
Furthermore, the field lacks consensus on evaluation standards, with substantial variability in criteria and protocols across studies \citep{howcroft-etal-2020-twenty}, complicating reliable and reproducible assessment.

\subsection{Instruction Following in LLMs}

Prompt engineering has emerged as an indispensable technique for guiding LLMs, enabling users to interact with generative AI systems through carefully designed instructions without modifying core model parameters \citep{sahoo2025systematicsurveypromptengineering, schulhoff2025promptreportsystematicsurvey}. Systematic surveys have categorized numerous distinct prompt engineering techniques based on their targeted functionalities \citep{sahoo2025systematicsurveypromptengineering}, though challenges persist including biases, factual inaccuracies, and interpretability gaps.

Research demonstrates that prompt formatting impacts LLM performance through various approaches. For example, the \textit{Chain-of-Thought (CoT)} technique prompts models to generate intermediate reasoning steps before delivering a final answer \citep{10.5555/3600270.3602070}. \textit{Optimization by PROmpting (OPRO)} employs an iterative process where the LLM generates new solutions from prompts containing previous solutions with their values, which are then evaluated and added to subsequent prompts \citep{yang2024large}. Studies also show that emotional intelligence can be leveraged through \textit{EmotionPrompt}, which combines original prompts with emotional stimuli to enhance performance \citep{li2023largelanguagemodelsunderstand}. Furthermore, the \textit{Rephrase and Respond} approach allows LLMs to rephrase and expand human-posed questions before providing responses, serving as an effective method for improving output quality \citep{deng2024rephraserespondletlarge}. These techniques collectively demonstrate that prompt quality influences response quality, with even simple formatting adjustments proving effective for performance improvement.

\section{Experimental Setup}

\paragraph{Datasets.}

We used two datasets covering different communication goals, languages and contexts: DialogSum~\citep{chen-etal-2021-dialogsum} and DECODA~\citep{favre-etal-2015-call}. 

DialogSum consists of social conversations in English and includes 12,460, 1,500 and 500 samples in its training, validation and test splits, respectively. It was used in a challenge~\citep{chen-etal-2022-dialogsum}, which favors comparison of results. 

DECODA~\citep{favre-etal-2015-call} is a call-center human-human spoken conversation corpus in French, collected from the RATP (Paris public transport authority) and mostly deals with customer inquiries and agent responses. The corpus was proposed as a pilot task at Multiling 2015~\citep{favre-etal-2015-call}. The test set used in this study contains 100 conversations with 212 synopses, i.e. references. 

\paragraph{Models and Parameters.}

We used OpenAI \chatgpt{} (\texttt{gpt-3.5-turbo}),\footnote{OpenAI GPT-3.5: \url{https://platform.openai.com/docs/models/gpt-3-5}, version gpt-3.5-turbo-0613} \gptfour{},\footnote{OpenAI GPT-4: \url{https://platform.openai.com/docs/models/gpt-4}, version gpt-4-0613} and \gptfouro{}\footnote{gpt-4o-2024-08-06: \url{https://platform.openai.com/docs/models\#gpt-4o}} APIs for our experiments.

\texttt{gpt-3.5-turbo} is optimized for conversational tasks while maintaining strong performance on standard completion tasks. The model handles inputs up to $4096$ tokens, compared to \gptfour{}'s $8192$-token limit and \gptfouro{}'s $128,000$-token context windows.
To ensure a stable output, we configured the temperature parameter at $0$ while maintaining the default values for all other parameters.

For comparison with task-specific pre-trained models, we also fine-tuned BART-large\footnote{\url{https://huggingface.co/facebook/bart-large}} on the DialogSum dataset and BARThez\footnote{\url{https://huggingface.co/moussaKam/barthez}} on the DECODA dataset.
The details of the training are presented in Appendix \ref{sec:appendix_bart_training}.

\paragraph{Summarization Prompt Design.}

\begin{table*}[!htb]
\scriptsize
\centering
        \resizebox{\textwidth}{!}{%
\begin{tabular}{p{3.5cm}|p{11.5cm}}
\toprule
\textbf{Settings} & \textbf{Prompts} \\
\hline
\textbf{WordLimit (WL)} \citep{laskar-etal-2023-systematic} & System: \textit{Write a summary with not more than X words. User: [Test Dialogue]} \\
\hline
\textbf{Human Guideline (HG)} for DialogSum & \textit{System: \textbf{Write a summary} based on the following criteria: the summary should (1) convey the most salient information of the dialogue and; (2) be brief (no longer than 20\% of the conversation length) and; (3) preserve important named entities within the conversation and; (4) be written from an observer perspective and; (5) be written in formal language. In addition, you should pay extra attention to the following aspects: 1) Tense Consistency: take the moment that the conversation occurs as the present time, and choose a proper tense to describe events before and after the ongoing conversation. 2) Discourse Relation: If summarized events hold important discourse relations, particularly causal relation, you should preserve the relations if they are also in the summary. 3) Emotion: you should explicitly describe important emotions related to events in the summary. 4) Intent Identification: Rather than merely summarizing the consequences of dialogues, you should also describe speakers’ intents in summaries, if they can be clearly identified. In addition to the above, you should use person tags to refer to different speakers if real names cannot be detected from the conversation. User: [Test Dialogue] } \\
\hline
\textbf{Human Guideline (HG)} for DECODA & \textit{System: \textbf{Write} a conversation-oriented summary in the form of a synopsis expressing both the customer's and the agent's points of view, and which should ideally report:
1. The main issues of the conversation: in call center conversations the main issues are the problems why the customer called; their identification constitutes the basis for classifying the call into several different classes of motivations for calling. 
2. The sub-issues in the conversation: when in the conversation any sub-issue occurs, it may be there both because it is introduced by the customer or by the agents.
3. The resolution of the call: i.e. if the customer’s problem was solved in that call (first-call resolution) or not. User: [Test Dialogue]} \\
\hline
\textbf{Human Guideline with Role (HGR)} & The same as the \textbf{Human Guideline (HG)} for DECODA but we changed the begin of instruction to \textit{\textbf{You are an annotator and are asked to write (dialogue summaries)} ...}. \\
\bottomrule
\end{tabular}
}
\caption{Prompts used for experiments. X is the average length of the reference summaries/synopses (DialogSum: X=20, DECODA: X=25). Prompts for DECODA have been translated from French.}
\label{tab:summarization_prompts}
\end{table*}

The different prompts used in our experiments are given in Table~\ref{tab:summarization_prompts}. 
The \texttt{WordLimit (WL)} prompt simply constrains the word length of the output to be less than $X$ words~\citep{laskar-etal-2023-systematic}. 
Another prompt we tested  was to provide a text only version of the full \texttt{Human Guideline (HG)}. For DialogSum, the annotation guideline was directly described in \citet{chen-etal-2021-dialogsum}. For DECODA, we contacted the authors \citep{favre-etal-2015-call} and obtained the guidelines used to write the synopses. 
In addition, to examine if giving the system a role helps, we proposed \texttt{Human Guideline with Role (HGR)}, which begins with \textit{"You are an annotator ..."}. 
We have also investigated a two-step iterative approach, called \texttt{HG(R)$\rightarrow$WL}, which first uses the summaries generated by \texttt{HG} or \texttt{HGR} as input, then re-injects them into the model to reduce the length of the output using the \texttt{WL} prompt. 

\paragraph{Evaluation Metrics.}
We reported the F1 scores of ROUGE-1/2/L \citep{lin-2004-rouge}, which assess the similarity between candidates and references based on the overlap of unigrams, bigrams, and the longest common sequence. We used a publicly available implementation.\footnote{\url{https://github.com/google-research/google-research/tree/master/rouge}}

We also reported the F1 scores of BERTScore \citep{Zhang*2020BERTScore:}, which measures the similarity between candidates and references at token level, using BERT's contextual embeddings.\footnote{\url{https://huggingface.co/spaces/evaluate-metric/bertscore}} For DialogSum, following \citet{chen-etal-2022-dialogsum}, we used \texttt{RoBERTa-large} \citep{Liu2019RoBERTaAR} as the backbone. For DECODA in French, we used the default multilingual model: \texttt{bert-base-multilingual-cased} \citep{devlin-etal-2019-bert}.

However, few metrics excel across all dimensions, leading recent studies to explore LLMs for evaluating text quality \citep{liu-etal-2023-g}.
Following this trend, we adopted an LLMs-as-judge approach, using \texttt{DeepSeek-R1} as the backbone for automatic evaluation on a subset of model predictions. 

\section{Results}

\begin{table*}[!htb]
\scriptsize
\centering
\begin{tabular}{p{1.8cm}|ccccc} 
\toprule
\textbf{DialogSum} & \textbf{R1} & \textbf{R2} & \textbf{RL} & \textbf{BS} & \textbf{Avg. Len} \\
\hline
\textbf{Human Ref.}  & 53.35  & 26.72 & 50.84  & 92.63 & 21.07 \\
\textbf{GoodBai} & {\bf 47.61} & {\bf 21.66} & 45.48  & {\bf 92.72} & 25.72 \\
\textbf{UoT} & 47.29 & 21.65   & {\bf 45.92}  & 92.26 & 27.05 \\
\textbf{BART-large} & 47.36 &  21.23  & 44.88  &  91.42  & 27.20 \\ 
\hline
\textbf{3.5-WL}  & 32.92 & \underline{12.45}  & 26.66  & 88.78  & 28.69 \\
\textbf{3.5-HG}  & 24.15 & 8.60  & 18.70  & 87.68 & 104.89 \\
\textbf{3.5-HGR}  & 26.43 & 9.27  & 20.50  & 88.16 & 88.85 \\ 
\rowcolor{Gray} 
\textbf{3.5-HG$\rightarrow$WL}  & 35.10 & 11.52  & 27.39  & 89.72 & 40.02 \\
\rowcolor{Gray}
\textbf{3.5-HGR$\rightarrow$WL}  & \underline{35.76} & 11.71  &  \underline{28.19}  & \underline{89.87} & 36.18 \\
\hline
\textbf{4-WL}  & 34.94 & \underline{12.38}  & \underline{28.99}  & 89.15  & 18.63 \\
\textbf{4-HGR}  & 26.25 & 8.01  & 19.86  & 88.44 & 86.78 \\
\rowcolor{Gray}
\textbf{4-HGR$\rightarrow$WL}  & \underline{35.42} & 9.58  &  28.20  & \underline{90.25} & 19.77 \\
\hline
\textbf{4o-WL} & \underline{35.34} & \underline{11.09} & \underline{28.83} & 89.33 & 18.62 \\
\textbf{4o-HGR} & 25.53 & 7.92 & 19.34 & 88.21 & 89.67 \\
\rowcolor{Gray}
\textbf{4o-HGR$\rightarrow$WL}  & 33.80 & 9.18 & 26.96  & \underline{89.83} & 20.23 \\
\bottomrule
\end{tabular}
\caption{Comparison of ROUGE-1/2/L, BERTScore, and average length for human references and summaries generated by different systems, including the top-performing models from the DialogSum challenge: GoodBai \citep{chen-etal-2022-dialogsum} and UoT \citep{lundberg-etal-2022-dialogue}.
Our results from \chatgpt{} (indicated as \texttt{3.5-}), \gptfour{} (indicated as \texttt{4-}) and \gptfouro{} (indicated as \texttt{4o-}) are presented alongside. The best results are in bold, and the top performances for \chatgpt{}, \gptfour{} and \gptfouro{} are underlined. Results from the two-step approach are highlighted in gray. Abbreviations: WL (WordLimit), HG (Human Guideline), HGR (Human Guideline with Role), BS (BERTScore).}
\label{tab: dialogsum_results}
\end{table*}

\begin{table*}[!htb]
\centering
\scriptsize
\begin{tabular}{p{1.8cm}|ccccc} 
\toprule
\textbf{DECODA} & \textbf{R1} & \textbf{R2} & \textbf{RL} & \textbf{BS} & \textbf{Avg. Len} \\
\hline
\textbf{Reference} & - & - & - & - & 25.40 \\
\textbf{BARThez}  & \textbf{35.42} & \textbf{16.96} & \textbf{29.41} & \textbf{74.94} & 18.53 \\
\hline
\textbf{3.5-WL} & \underline{33.21} & \underline{12.53} & \underline{24.74} & \underline{72.54} & 37.90 \\ 
\textbf{3.5-HG} & 18.42 & 7.18 & 13.68 & 67.72 & 162.83 \\
\textbf{3.5-HGR} & 16.72 & 6.73 & 12.66 & 66.93 & 190.87 \\
\rowcolor{Gray}
\textbf{3.5-HG$\rightarrow$WL}  & 31.61 & 11.00  & 23.34  & 71.77 & 46.30 \\
\rowcolor{Gray}
\textbf{3.5-HGR$\rightarrow$WL}  & 32.74 & 12.22  & 24.09  & 72.30 & 41.76 \\
\hline
\textbf{4-WL} & \underline{35.23} & \underline{13.54} & \underline{27.73} & \underline{73.23} & 23.81 \\ 
\textbf{4-HGR} & 20.71 & 8.29 & 15.12 & 69.04 & 136.53 \\
\rowcolor{Gray}
\textbf{4-HGR$\rightarrow$WL}  & 32.12 & 11.17  & 24.82  & 72.30 & 26.01 \\
\hline
\textbf{4o-WL} &  \underline{34.86} & \underline{13.05} & \underline{27.04} & \underline{73.59} & 22.73 \\
\textbf{4o-HGR} & 19.35 & 7.73 & 13.94 & 68.57 & 160.73 \\
\rowcolor{Gray}
\textbf{4o-HGR$\rightarrow$WL}  & 31.43 & 10.40 & 23.67  & 72.42 & 24.17 \\
\bottomrule
\end{tabular}
\caption{Comparison of ROUGE-1/2/L, BERTScore, and average length for references and the state-of-the-art fine-tuned model \citep{zhou-etal-2022-effectiveness} against \chatgpt{} (indicated as \texttt{3.5-}), \gptfour{} (indicated as \texttt{4-)} and \gptfouro{} (indicated as \texttt{4o-}) on the DECODA dataset.
The best overall results are in bold, and the top performances for \chatgpt{}, \gptfour{} and \gptfouro{} are underlined. Results from the two-step approach are highlighted in gray. 
Abbreviation: WL (WordLimit), HG (Human Guideline), HGR (Human Guideline with Role), BS (BERTScore).
} 
\label{table:decoda_results}
\end{table*}

\subsection{Quantitative Results}

\paragraph{DialogSum}
Table~\ref{tab: dialogsum_results} presents the automatic evaluation results on the test set.
While the \chatgpt{} model with the \texttt{WordLimit} prompt underperforms state-of-the-art pre-trained models by approximately 15 (R1), 10 (R2), and 20 (RL) points, BERTScore indicates only a minor semantic discrepancy. The \gptfour{} model with the same prompt shows a slight overall improvement over \chatgpt{}, except on R2. Furthermore, \gptfouro{} performs comparably to \gptfour{}, with the \texttt{WL} prompt yielding superior results on most metrics, though not on BERTScore.

Although the DialogSum guideline states \textit{"be brief (no longer than 20\% of the conversation length)"}, the results of the \texttt{HG} prompt indicate that instructions tailored for human annotators may not be suitable when directly used as instructions to generate reference-like summaries with \chatgpt{}, as responses tend to be longer, leading to lower ROUGE and BERTScore values. This effect could have been reduced by specifying the role of the annotator in the prompt (\texttt{HGR}), which consistently improved performance with reference to \texttt{HG} across all measures.

Regarding the two-step prompt approach \texttt{HGR$\rightarrow$WL}, which incorporates human annotation instructions as a first step, followed by applying a second prompt to limit word length, the performance aligned with those obtained with a single \texttt{WordLimit} prompt, with even higher R1 and RL scores, suggesting that the GPT model's word usage is more akin to the references, maintaining the primary content and logical order, albeit with some variability in language usage.

In terms of average length, \gptfour{} and \gptfouro{} better followed the \texttt{WordLimit} prompt in obtaining a summary whose length is closest to the reference compared to \chatgpt{}.

\paragraph{DECODA}

Results in Table~\ref{table:decoda_results} show that the \texttt{WordLimit} prompt produces the best outcomes for both GPT models. However, their performance still falls short of the pre-trained model across all metrics. The difference is less pronounced in the BERTScore than in the ROUGE scores. 

When using human guidelines (\texttt{HG(R)}), results diverged from those of the other prompts, resulting in reduced ROUGE and BERTScore. For \chatgpt{}, \texttt{HGR} showed lower results than \texttt{HG}.
The two-step prompting \texttt{HG(R)$\rightarrow$WL} produced shorter final summaries than \texttt{HG(R)}, resulting in higher scores across various measures, although they remain slightly lower than those obtained with the simple \texttt{WordLimit} prompt.

When comparing the GPT models, all three models performed better with the \texttt{WordLimit} prompt, while yielding similar results with the iterative prompt (\texttt{HGR$\rightarrow$WL}).
In terms of average length, the \texttt{WordLimit} prompt consistently gave the shortest summaries for all GPT models. \gptfour{} and \gptfouro{} followed better the \texttt{WordLimit} prompt in obtaining a summary of less than 25 words than \chatgpt{}, which consistently exceeded the limit.

\paragraph{Using LLM-as-judge}

Using \texttt{DeepSeek-R1}\footnote{\url{https://api-docs.deepseek.com/guides/reasoning_model}} as our evaluation backbone, we evaluated the generated summaries against the human summarization guideline criteria (see Appendix \ref{appendix_sec:llm_as_judge} for details and prompts).
We evaluated a subset of 20 samples from the DECODA dataset, with results shown in Table \ref{tab:llm_as_judge_r1}. 
To investigate the impact of length on quality, we include an additional column reporting summary length in our analysis.

However, we observed that even when the prompt explicitly provided guidelines for judging summaries and specified that only a score should be returned, many evaluation results contained not only scores but also explanatory justifications (sometimes in French and sometimes in English).

The GPT models' generated summaries achieved strong scores across all four criteria. The best results for \textit{Faithfulness}, \textit{Main Issues}, and \textit{Resolution} came from \texttt{4-HGR$\rightarrow$WL}, \texttt{3.5-WL} and \texttt{3.5-HGR$\rightarrow$WL} respectively, while the \texttt{Reference} performed best on \textit{Sub-Issues}. Considering the average scores across all criteria, \texttt{4-HGR$\rightarrow$WL} achieved the highest values, while \texttt{BARThez} achieved the lowest.

\begin{table*}[!htb]
    \centering
    \footnotesize
        \resizebox{\textwidth}{!}{%
    \begin{tabular}{lcccccc}
    \toprule
     & \textbf{Faith. (VN)} & \textbf{Main Iss. (VN)}  & \textbf{Sub-Iss. (VN)}  & \textbf{Resol. (VN)}  & \textbf{Avg. (VN)}  & \textbf{Length} \\
    \midrule
    \textbf{Reference}          & 3.65$\pm$1.09 (20)  & 4.35$\pm$1.09 (20) & \textbf{3.85$\pm$1.27} (20)  & 3.15$\pm$1.81 (20) & 3.75$\pm$1.32 (20) & 24.85$\pm$17.19\\
    \textbf{3.5-WL}      & 4.00$\pm$0.86 (20)  & \textbf{4.60$\pm$0.75} (20)  & 3.60$\pm$1.10 (20)  & 3.35$\pm$1.73 (20) & 3.89$\pm$1.11 (20) & 44.00$\pm$23.86\\
    \textbf{3.5-HGR$\rightarrow$WL}    & 3.20$\pm$1.47 (20)  & 4.00$\pm$1.30 (20)  & 3.75$\pm$1.21 (20)  & \textbf{3.95$\pm$1.15} (20) & 3.73$\pm$1.28 (20) & 38.40$\pm$10.18\\
    \textbf{4-WL}        & 4.00$\pm$1.12 (20)  & 4.40$\pm$0.82 (20)  & 3.42$\pm$1.39 (19)$^\dag$  & 2.45$\pm$1.70 (20) & 3.57$\pm$1.26 (20) & 23.35$\pm$~3.08\\
    \textbf{4-HGR$\rightarrow$WL}      & \textbf{4.15$\pm$0.93} (20)  & 4.45$\pm$0.76 (20) & 3.80$\pm$0.83 (20)  & 3.50$\pm$1.54 (20) & \textbf{3.98$\pm$1.02} (20) & 25.20$\pm$~3.30 \\
    \textbf{BARThez}           & 1.70$\pm$1.13 (20)  & 2.00$\pm$1.65 (20)  & 1.42$\pm$0.84 (19)$^\ddag$ & 1.60$\pm$1.27 (20) & 1.68$\pm$1.22 (20) & 18.05$\pm$~5.04 \\
    \bottomrule
\end{tabular}
        }
    \vspace{0.2cm}
    \footnotesize{
    $^\dag$One sample in 4-WL was excluded due to "N/A". \\
    $^\ddag$One sample in BARThez was excluded due to an error in returning the score.
    }
    \caption{Results of using \texttt{DeepSeek-R1} for LLM-as-judge evaluation. Results are presented as Mean $\pm$ Std. (VN) indicates the "validated number" of samples (out of 20) used to calculate the score for that criterion. \textit{Length} indicates summary word count.}
    \label{tab:llm_as_judge_r1}
\end{table*}

\subsection{Quantitative Results by Variation}

\begin{table*}[!htb]
    \centering
    \footnotesize
    \begin{tabular}{lcccc}
    \toprule
    & ROUGE-1 & ROUGE-2 & ROUGE-L & BERTScore \\
    \midrule
    Human Variance & 53.37 (16.75) & 26.73 (20.57) & 44.99 (18.11) & 92.86 (2.68) \\
    \midrule
    4-WL & 34.96 & \textbf{12.35} & \textbf{28.99} & 89.15 \\
    4-HGR & 26.24 & \textbf{8.01} & 19.86 & 88.44 \\
    4-HGR$\rightarrow$WL & 35.43 & \textbf{9.58} & \textbf{28.18} & \textbf{90.25} \\
    \bottomrule
\end{tabular}

    \caption{Automatic metric results on the DialogSum test set. The \textit{Human Variance} row reports the mean (std) of \textit{pairwise comparisons} between all three human reference summaries.
    Model scores in \textbf{bold} fall within one standard deviation of this human variance mean.}
    \label{tab:variance_results}
\end{table*}

For the DialogSum dataset, each dialogue is annotated by three annotators, resulting in three reference summaries per dialogue. Despite the shared annotation guideline, human annotators can exhibit variability in writing style. To evaluate whether GPT models follow human summarization guidelines, we compare the models' scores to the variance observed in the reference summaries for each metric (ROUGE-1/2/L and BERTScore). This approach determines whether the model-generated summaries fall within a reasonable range of human-annotated references, based on the mean and standard deviation of the reference scores.

We evaluated predictions from \texttt{4-WL}, \texttt{4-HGR}, and \texttt{4-HGR$\rightarrow$WL}, with results presented in Table \ref{tab:variance_results}. The first row -- Mean (std) -- represents the variance among reference summaries from three annotators. Our findings show that \texttt{4-WL} predictions fall within the reference variance for R2 and RL, \texttt{4-HGR} only for R2, and \texttt{4-HGR$\rightarrow$WL} for R2, RL, and BERTScore. These results demonstrate that, similar to how different annotators exhibit varying styles in writing references, the generated summaries align with the variance distribution of human annotators for certain metrics.

\subsection{Manual Error Analysis}
\label{subsec:manual_error_analysis}

We examined the data points where GPT-generated summaries exhibit the greatest discrepancy, characterized by low ROUGE scores but high BERTScore values. 
For comparison, we selected predictions from two models: \texttt{4-WL} (simple \textit{WordLimit} prompt) and \texttt{4-HGR$\rightarrow$WL}, which incorporates human summarization guidelines as an intermediate step.
For DialogSum, we show top discrepancies from \texttt{4-WL} (Table~\ref{tab:discrepancy_dialogsum_4_WL}) and \texttt{4-HGR$\rightarrow$WL} (Table \ref{tab:discrepancy_dialogsum_4_HGR_WL}). 
For DECODA, we present \texttt{4-WL} results (Table~\ref{tab:discrepancy_decoda_4_WL_translation_en}) and \texttt{4-HGR$\rightarrow$WL} (Table \ref{tab:discrepancy_decoda_4_HGR_WL_translation_en}), both translated from French.

\begin{table*}[!htb]
    \scriptsize
            \resizebox{\textwidth}{!}{%
    \begin{tabular}{lp{8.3cm}p{3.5cm}llll} 
\toprule
\textbf{ID}  & \textbf{References}    & \textbf{Prediction}    & \textbf{RL}  & \textbf{BS} & \textbf{Diff.} \\
\midrule
16  & {[}"\#Person2\# at first thinks \#Person1\#'s behaviour cruel but finally   joins \#Person1\#.", "\#Person1\# plans on playing a trick to others.   \#Person2\# thinks it's cruel at first but then joins.", "\#Person1\#   is about to make a prank. \#Person2\# thinks it's cruel at first but then   joins."{]}   & Two friends engage in a potentially offensive prank, causing a   disagreement about their choice of fun.                         & 6.00           & 86.09             & 80.09      \\
\midrule
24  & {[}"\#Person1\# is interviewing \#Person2\# about \#Person2\#'s ability and   previous experience.", "\#Person1\# asks \#Person2\# about \#Person2\#'s   capacities and past working experience during an interview.",   "\#Person1\# asks \#Person2\# about \#Person2\#'s qualification for the   job."{]}    & Person 2, an Automobile Designing graduate, has experience in   understanding mechanical strength and corrosion resistance.      & 4.86         & 83.67             & 78.81      \\
\midrule
255 & {[}"\#Person1\# and \#Person2\# are conjecturing who will get sacked based   on other employees' performance and find that it might be themselves.   \#Person1\# is so anxious as \#Person1\# needs to raise family but \#Person2\# will   be happy to leave.", "\#Person1\# and \#Person2\# talk about the   company's upcoming downsizing and discuss who will get sacked. \#Person1\# is   worried but \#Person2\# doesn't care.", "\#Person1\# and \#Person2\# have   rounds of conjectures of whom shall be fired due to the company's cutback and   comes to the conclusion that they both have the biggest possibility. It   drives \#Person1\# crazy but doesn't worry \#Person2\#."{]}   & Employees speculate about potential layoffs at their company, considering   various colleagues' relationships with the boss.     & 7.15         & 85.90              & 78.76      \\
\midrule
48  & {[}'\#Person1\# and \#Person2\# are talking about the low temperature at night,   although spring has come.', '\#Person1\# and \#Person2\# agree that it still felt   very cold in spring.', '\#Person1\# and \#Person2\# talk about the weather and   how to keep warm.'{]}  & Despite spring's arrival, cold nights persist, causing discomfort indoors   and outdoors.                                        & 8.02         & 86.53             & 78.50       \\
\midrule
67  & {[}'\#Person1\# is driving \#Person2\# to an inn. They talk about their   careers, ages, and where they was born.', '\#Person1\# drives \#Person2\# to an   inn and they have a talk. \#Person2\# is 26 and had a business trip to China.   \#Person1\# is 40 years old American.', '\#Person1\# drives \#Person2\# from the   airport to an inn and they have a casual talk about themselves.'{]}  & A Mexican businesswoman, returning from China, converses with her   American-Colombian taxi driver.                              & 7.22         & 84.96             & 77.74      \\
\bottomrule
\end{tabular}
    }
    \caption{\label{tab:discrepancy_dialogsum_4_WL} Data points where GPT-generated summaries (\texttt{4-WL}) show the greatest discrepancy in the DialogSum corpus, with low ROUGE-L but high BERTScore values. RL and BS are the average scores of three references.} 
\end{table*}

\begin{table*}[htb]
\scriptsize
        \resizebox{\textwidth}{!}{%
    \begin{tabular}{lp{5.8cm}p{6cm}lll}
\toprule
\textbf{ID}  & \textbf{Reference}    & \textbf{Prediction}    & \textbf{RL}  & \textbf{BS} & \textbf{Diff.}  \\
\midrule
10  & Request for information about a large account order. Transferred to the relevant service.  
    & A seminar organizer is seeking transport cards for their international guests from the RATP.  
    & 0      & 71.13    & 71.13      \\
\midrule
110 & Request for information following the procedure after receiving a fine notice. Transferred to the relevant service.  
    & A customer calls customer service to understand why she has to pay five euros after contesting and justifying a bus fine.  
    & 8.33 & 70.86    & 62.53      \\
\midrule
107 & request for a refund procedure following a ticket purchase due to the late delivery of the ImagineR pass, with a waiting period of over 3 weeks. Refund possible, contact ImagineR by phone or email.  
    & A customer calls to request a refund for an Orange card purchased while waiting for her Imagine+R card, which arrived late.  
    & 8.00   & 69.35    & 61.35      \\
\midrule
32  & T2 circulation  
    & A user inquires about the launch of the T+two line from Porte de Versailles to Val d'Issy.  
    & 8.70  & 69.85    & 61.15   \\
\midrule
29  & Request for information on purchasing tickets for school groups. Transferred to the relevant service.  
    & A representative of a departmental association is looking to organize transport for more than 100 classes using public transportation.  
    & 10.26 & 71.05    & 60.80    \\
\bottomrule
\end{tabular}

    }
\caption{\label{tab:discrepancy_decoda_4_WL_translation_en} Data points (translated version) where GPT-generated summaries (\texttt{4-WL}) show the greatest discrepancy in the DECODA corpus, with low ROUGE-L but high BERTScore values.}
\end{table*}

\paragraph{4-WL} We first analyze the summaries generated by \texttt{4-WL}.
In DialogSum examples with the greatest discrepancies, the references preserve specific mentions such as \textit{\#Person1\#} and \textit{\#Person2\#}, retaining named entities while summarizing the conversation. In contrast, \gptfour{} predictions often provide a higher-level generalization, using generic terms like "two friends" or "employees" instead of specific tags. This indicates that GPT-generated summaries do not meet the following guideline: "\textbf{3) Preserve important named entities within the conversation}" when using a simplified prompt. 

In addition, \gptfour{} predictions do not match the references concerning the guideline emphasizing "\textbf{Intent Identification}".\footnote{"Rather than merely summarizing the consequences of dialogues, you should also describe speakers intents in summaries, if they can be clearly identified. In addition to the above, you should use person tags to refer to different speakers if real names cannot be detected from the conversation."}
For instance, in the second example, the prediction references only \textit{Person 2}, omitting the intent of \textit{Person 1}, who is interviewing the other individual. 
In the fourth example, neither \textit{Person 1} nor \textit{Person 2} is mentioned, and their intents are absent, as the prediction reduces the dialogue to a general remark about the weather. 
In the fifth example, the prediction fails to follow the "\textbf{use person tags}" rule, summarizing the conversation by identifying the interlocutors as \textit{"A Mexican businesswoman"} and \textit{"her American-Colombian taxi driver"}. 
These variations probably influenced the quantitative results, as evidenced by the disparity between high BERTScore values, which indicate strong semantic similarity, and low ROUGE-L scores, which reflect minimal lexical overlap. This contrast highlights the tendency of GPT-generated summaries using a simple length-constrained prompt to capture the essential meaning and key information of the conversation while diverging from the reference summaries in their specific wording and structural composition.

For the DECODA dataset, the discrepancy may stem from the distinct characteristics of the reference synopses. These summaries focus strictly on issues and resolutions, whereas GPT-generated predictions often adopt a more \textbf{individual-centered perspective}, using terms such as \textit{"un organisateur (an organizer)"}, \textit{"une cliente (a client)"}, and \textit{"un représentant (a representative)"}, etc. This difference results in relatively high BERTScore values, reflecting strong semantic similarity, but low ROUGE-L scores due to minimal word overlap with the references. Additionally, reference synopses are often extremely brief and sometimes incomplete sentences, as seen in the fourth example: \textit{"circulation du T2"} (T2 circulation). Furthermore, all samples except the fourth fail to conclude the resolution in the generated summaries.

\paragraph{4-HGR$\rightarrow$WL} Next, we examined the generated summaries of \texttt{4-HGR$\rightarrow$WL}. On the DialogSum dataset, we observe better retention of named entities and personal tags (e.g., \textit{Person 1}) compared to \texttt{4-WL}, indicating that the intermediate \texttt{HGR} step helps the model to follow specific rules from the human summarization guidelines. However, the issue of overlooking "\textbf{Intent Identification}" persists, as prediction often focus on one individual's intent while omitting the other's, whereas references typically indicate both interlocutors' intents.

For DECODA, similar to the \texttt{4-WL} output, most generated summaries employ an individual-centered perspective that creates divergence in $n$-gram overlap and semantic similarity. Although reference summaries maintain conciseness, model predictions tend toward detail while frequently omitting critical "\textbf{Resolution}" elements. For example, the first sample omits "\textit{Transfer to the relevant department}" despite this being an explicit requirement in the summarization guidelines.

In summary, comparing cases with the greatest discrepancy (low ROUGE-L but high BERTScore) between \texttt{4-WL} and \texttt{4-HGR$\rightarrow$WL}, the latter adheres better to guidelines for named entities and personal tags. However, both models share similar shortcomings: in DialogSum, they overlook "Intent Identification"; in DECODA, they use an individual-centered perspective and omit "Resolution".

\subsection{Human Evaluation}

\begin{figure*}[htbp]
\begin{center}
\includegraphics[width=.9\textwidth]{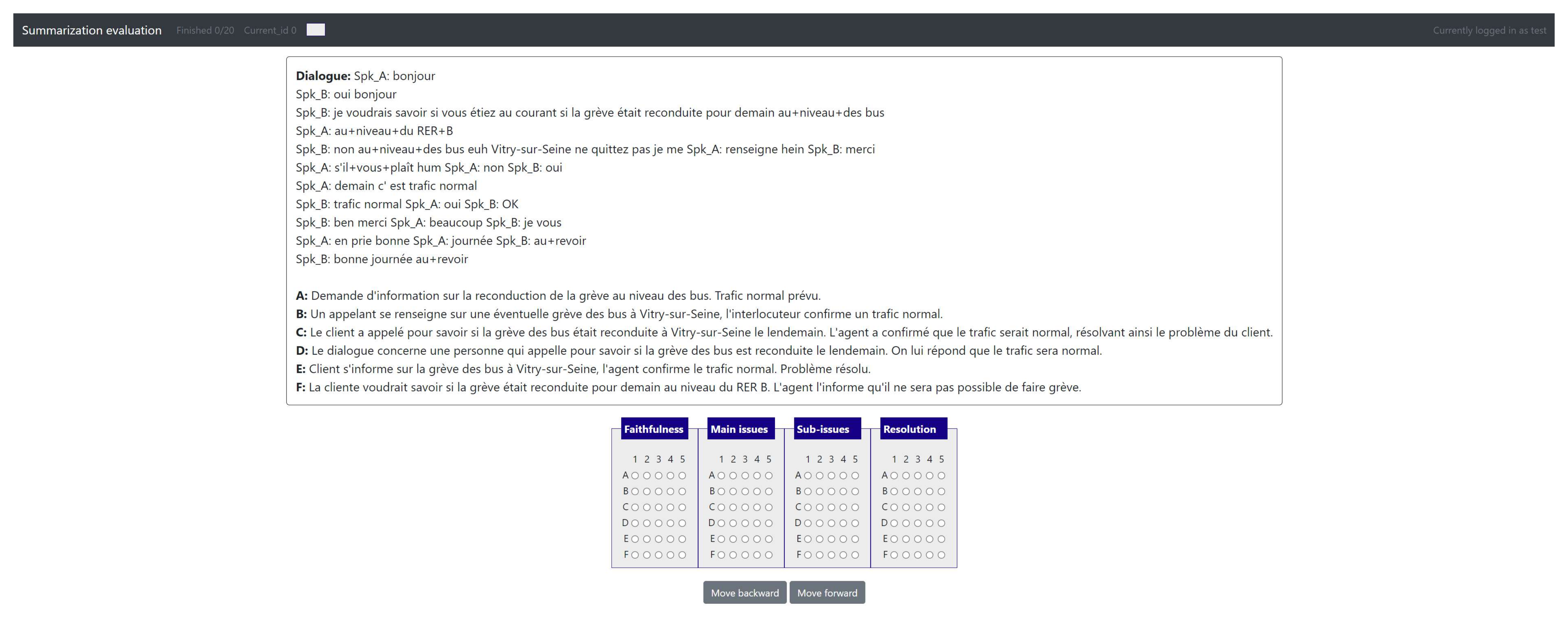}
\caption{Screenshot of the interface used for evaluating summaries in the DECODA summarization task.}
\label{fig:human_eval_interface}
\end{center}
\end{figure*}

Due to the misalignment between automatic measures and human judgment of LLM outputs, as evidenced in news summarization experiments with GPT-3 \citep{goyal2023news}, we set up a human evaluation. 
To determine if the generated summaries meet predefined instructions, we conducted a human evaluation using criteria derived from the annotation guidelines provided to human annotators for writing reference summaries. We started with the DECODA dataset, as its task-oriented summarization guidelines are more easily adaptable into evaluation criteria.
We selected the 10 shortest and 10 longest dialogues, assessing six summaries for each dialogue, including references and predictions from various models: \texttt{BARThez}, \texttt{3.5-WL}, \texttt{3.5-HGR$\rightarrow$WL}, \texttt{4-WL}, and \texttt{4-HGR$\rightarrow$WL}. 

\paragraph{Annotation interface}

For our human evaluation task, we adapted the POTATO annotation tool \citep{pei-etal-2022-potato}. The evaluation interface, shown in Figure \ref{fig:human_eval_interface}, features a dialogue at the top, accompanied by six different summaries, including references and model outputs from \texttt{BARThez}, \texttt{3.5-WL}, \texttt{3.5-HGR$\rightarrow$WL}, \texttt{4-WL}, and \texttt{4-HGR$\rightarrow$WL}. To minimize bias, we randomized the order of summaries, assigning them names A, B, C, D, E, and F. 

There are four blocks representing the different evaluation criteria.
In each block, annotators interact with the interface by clicking on each of the six summaries, assigning a score from 1 to 5 to each summary based on its compliance with the specific criterion, 5 being the best.

Three native French speakers took part in the evaluation, all graduate students. We presented them with a PDF evaluation guide and explained that the annotations they made would be used for analysis.

\paragraph{Criteria} 
Annotators assessed each summary based on four criteria: \textit{Faithfulness}, \textit{Main Issues}, \textit{Sub-Issues}, and \textit{Resolution}. The later three criteria are based on human summarization guidelines, while \textit{Faithfulness} was added to evaluate the accuracy of the facts presented in relation to the dialogues. \textit{Sub-Issues} was only evaluated if at least one sub-issue was present in the dialogue.

The evaluation criteria are described below. In the evaluation guide, we also provide detailed explanations for each score of criterion.

\begin{itemize}[noitemsep,topsep=0pt,parsep=0pt,partopsep=0pt]
    \item \textbf{Faithfulness}: the summary must respect the dialogue at the level of factual information.
    \item \textbf{Main issues}: in call center conversations the main issues are the problems why the customer called; their identification constitutes the basis for classifying the call into several different classes of motivations for calling.
    \item \textbf{Sub-issues}: when in the conversation any sub-issue occurs, it may be there both because it is introduced by the customer or by the agents.
    \item \textbf{Resolution}: i.e. if the customer's problem was solved in that call or not. 
\end{itemize}

To provide context for human evaluation scores, we include a \textit{Length} column indicating the word count of summaries.

\paragraph{Results} 

\begin{table*}[!htb]
\footnotesize
\centering
  \begin{tabular}{lcccccc}
\toprule
\textbf{} & \textbf{Faithfulness} & \textbf{Main Issues} & \textbf{Sub-issues} & \textbf{Resolution} & \textbf{Average} & \textbf{Length} \\
\midrule
Reference & 3.72$\pm$1.12 & 4.15$\pm$1.07 & 3.22$\pm$1.86 & 3.25$\pm$1.60 & 3.68$\pm$1.36 & 24.85$\pm$17.19 \\
3.5-WL & 4.02$\pm$1.17 & 4.30$\pm$1.18 & 2.67$\pm$1.73 & 4.18$\pm$1.28 & 4.10$\pm$1.28 & 44.00$\pm$23.86 \\
3.5-HGR$\rightarrow$WL & 3.97$\pm$1.15 & \textbf{4.47$\pm$0.89} & \textbf{3.67$\pm$1.73} & \textbf{4.20$\pm$1.13} & \textbf{4.19$\pm$1.12} & 38.40$\pm$10.18 \\
4-WL & \textbf{4.13$\pm$0.96} & 4.37$\pm$0.80 & 1.44$\pm$1.33 & 2.70$\pm$1.79 & 3.62$\pm$1.53 & 23.35$\pm$~3.08 \\
4-HGR$\rightarrow$WL & 3.93$\pm$1.02 & 4.30$\pm$0.96 & 2.67$\pm$1.66 & 4.05$\pm$1.27 & 4.03$\pm$1.16 & 25.20$\pm$~3.30 \\
BARThez & 2.17$\pm$1.43 & 2.32$\pm$1.57 & 1.00$\pm$0.0 & 2.22$\pm$1.54 & 2.17$\pm$1.49 & \textbf{18.05$\pm$~5.04} \\
\bottomrule
\end{tabular}
  \caption{
Mean ($\pm$ std) scores for \textit{Faithfulness}, \textit{Main issues}, \textit{Sub-issues}, \textit{Resolution}, and \textit{Overall} of each summary assessed by three native human evaluators on the 10 shortest and 10 longest dialogues in the DECODA dataset. The best results for each criterion are shown in bold.}
\label{tab:human_eval_results_part}
\end{table*}

Table~\ref{tab:human_eval_results_part} presents the human evaluation results for the 20 evaluated dialogues across four criteria and their overall scores. A comprehensive comparison, including results for all dialogues, the 10 shortest, and the 10 longest dialogues, is provided in Table~\ref{tab:human_eval_results_full} in the Appendix.
Overall, \texttt{3.5-HGR$\rightarrow$WL} achieved the highest score of 4.19, though \texttt{4-WL} slightly excelled in \textit{Faithfulness}. 
For the 10 shortest dialogues, \texttt{3.5-HGR$\rightarrow$WL} led with an overall score of 4.44, while \texttt{3.5-WL} had the highest overall score at 4.08 for the 10 longest dialogues.

The \textit{Sub-issues} criterion proved challenging to analyze for two primary reasons: sub-issues were not present in all dialogues, and evaluators demonstrated significant variability in their assessments. Specifically, one annotator identified sub-issues in five dialogues, another in three, and a third in only one dialogue.
In addition, annotators generally preferred the outputs from GPT models over the reference summaries, with \texttt{BARThez} being the least favored overall. However, \texttt{4-WL} received the lowest scores specifically for the \textit{Resolution} criterion among the 10 longest dialogues.

The annotators' preference for GPT-generated summaries may stem from the models' use of more natural and fluid language. In contrast, the reference synopses in the DECODA dataset are highly specific and concise.
For example, they are often short and to the point, such as: \textit{"Request for train timetables from Maux station to Gare de l'Est station at a given time"}.
In some cases, the reference synopses are not even complete sentences. This stylistic contrast likely explains the annotators' tendency to favor the more cohesive and natural language of the GPT-generated outputs.

\paragraph{Inter-Annotator Agreement (IAA)}

\begin{table*}[!htb]
\centering
\footnotesize
\begin{tabular}{lccc}
\toprule
 & \textbf{A1-A2} & \textbf{A1-A3} & \textbf{A2-A3} \\
\midrule
\textbf{Faithfulness} & 0.422 & 0.273 & 0.315 \\
\textbf{Main issues}  & 0.363 & 0.290  & 0.398 \\
\textbf{Sub-issues}   & 0.140  & 0.445 & -0.175 \\
\textbf{Resolution}   & 0.546 & 0.503 & 0.463 \\
\bottomrule
\end{tabular}
\caption{Inter-annotator agreement (IAA) scores among three annotators for dialogue summary evaluation on the DECODA dataset.}
\label{tab:decoda_evaluation_iaa}
\end{table*}

We calculated inter-annotator agreement (IAA) scores for the three annotators using Cohen's kappa, the results are presented in Table \ref{tab:decoda_evaluation_iaa}. 
The analysis reveals stronger agreement on the \textit{Resolution} criterion, whereas annotations for \textit{Sub-issues} showed considerable variation.

\section{Discussion and Conclusion}

Human evaluation results show that GPT models (\chatgpt{}, \gptfour{} and \gptfouro{}) can follow human summarization guidelines to some extent. Their summaries were preferred over task-specific pre-trained models and even outperformed reference summaries. The longer outputs of GPT models, compared to BARThez, likely contribute to more comprehensive coverage of issues and resolutions, enhancing faithfulness to dialogues.
However, GPT models underperform task-specific pre-trained models on both ROUGE and BERTScores metrics. This discrepancy between automatic metrics and human judgments aligns with previous findings, reinforcing (1) the continued necessity of human evaluation for text generation tasks, and (2) the critical need for developing better-aligned automatic evaluation metrics.

The results also indicate that GPT models demonstrate some proficiency in adhering to human guidelines. In the overall evaluation using human guidelines, the \texttt{HGR$\rightarrow$WL} approach tends to produces better results than the simpler \texttt{WordLimit} prompt. 
Interestingly, \gptfour{} does not consistently outperform \chatgpt{}; for instance, \texttt{4-WL} only surpasses \texttt{3.5-WL} on the 10 shortest dialogues. Nevertheless, \gptfour{} exhibits better adherence to the specified word length constraints.

Our findings also highlight the inherent subjectivity of summary quality assessment, as evidenced by the human evaluators' preference for GPT-generated summaries over reference texts, which is partly attributable to stylistic differences.

By analyzing data points where GPT-generated summaries (\texttt{4-WL} and \texttt{4-HGR$\rightarrow$WL}) exhibit significant discrepancies -- characterized by low ROUGE scores but high BERTScore values -- we found that the models effectively summarize key dialogue information and are semantically similar to the references. However, they sometimes neglect specific rules: in DialogSum, they miss named entities, intent identification, and personal tags; in DECODA, some summaries focus on individual locutors' perspectives and fail to present resolutions. The intermediate step of using \texttt{HGR} helped the model better adhere to the specific rules of human summarization guidelines, such as the use of named entities and personal tags in DialogSum.

For future work, we plan to develop LLM-based metrics that evaluate summaries against task-specific guidelines, with a focus on improving response stability and alignment with human judgment.
Furthermore, we will investigate how LLMs follow different summarization guidelines, including those with lexical perturbations, to interpret their decision-making processes and operational mechanisms.

\section*{Limitations}
In terms of model selection, our analysis was limited to GPT models: \chatgpt{}, \gptfour{} and \gptfouro{}. Future studies could benefit from a broader exploration of LLMs, encompassing models with varying parameter sizes and accessibility.

\paragraph*{Supplementary Materials Availability Statement:}
\textbf{Code Resources}: The full experimental codebase is available at \url{https://github.com/yongxin2020/LLM-Sum-Guidelines}, including the LLM-as-judge framework in the \texttt{Guideline-Eval/} directory.

\textbf{Datasets}: The DialogSum dataset is publicly available at \url{https://github.com/cylnlp/dialogsum}, while the DECODA dataset can be obtained upon request from \url{https://pageperso.lis-lab.fr/benoit.favre/cccs/}.

\textbf{Generated Outputs}: All model predictions are stored in \texttt{decoda/output/} and \texttt{dialogsum/output/} directories, with two-step prompting results in \texttt{*/twoSteps/} subdirectories. BART baseline outputs are provided in \texttt{sota\_barthez\_predictions.txt} and \texttt{bart\_large\_summaries.txt}.

\textbf{Human Evaluation}: The annotation protocol is documented in \texttt{HumanEval\_Guideline\_DECODA.pdf}, with raw annotations available in \texttt{results/human\_annotations\_decoda/} and analysis scripts in \texttt{human\_eval\_scores.py}.

\section*{Acknowledgments}
This research was supported by the Banque Publique d’Investissement (BPI) under grant agreement THERADIA and was partially supported by MIAI@Grenoble-Alpes (ANR-19-P3IA-0003 and ANR-23-IACL-0006). We thank the anonymous reviewers for their insightful comments, as well as the annotators who participated in the human evaluation process.

\bibliographystyle{unsrtnat}
\bibliography{references}  

\appendix

\section{Experiment details: BART-based models}
\label{sec:appendix_bart_training}

\paragraph{DialogSum}
We fine-tuned the BART-Large model\footnote{\url{https://huggingface.co/facebook/bart-large}} on the DialogSum dataset using the following hyperparameters: a learning rate of $5 \times 10^{-5}$ for 15 epochs, a training batch size of 2, and an evaluation batch size of 8. The maximum source and target lengths were set to 1024 and 128 tokens, respectively, with a length penalty of 1.0. The experiments were conducted on an NVIDIA Quadro RTX 6000 GPU, with each run taking approximately 2.5 hours.
    
\paragraph{DECODA}

We fine-tuned the BARThez model\footnote{\url{https://huggingface.co/moussaKam/barthez}} on the preprocessed DECODA dataset. 
The model, which uses a base architecture with 6 encoder and 6 decoder layers, was trained for 10 epochs, and the checkpoint with the lowest loss on the development set was saved.
We employed the default parameters for summarization tasks: an initial learning rate of $5 \times 10^{-5}$, a training and evaluation batch size of 4, and a seed of 42. 
The Adam optimizer with a linear learning rate scheduler was used. 
All experiments were run on an NVIDIA Quadro RTX 8000 48GB GPU, with each training run taking approximately 25 minutes.

\section{LLM-as-judge}
\label{appendix_sec:llm_as_judge}

For the LLM-as-judge approach, we adapted the G-Eval framework \citep{liu-etal-2023-g}, modifying its criteria to align with DECODA's dialogue summarization guidelines. Specifically, we evaluated summaries based on \textit{Faithfulness} (Figure \ref{fig:prompt_faithfulness}), \textit{Main Issues} (Figure \ref{fig:prompt_main_issues}), \textit{Sub-Issues} (Figure \ref{fig:prompt_sub_issues}), and \textit{Resolution} (Figure \ref{fig:prompt_resolution}). 

Each criterion was assessed via a distinct, specially designed prompt (see corresponding figures).

\begin{figure*}[!ht]
    \input{Tables/prompts/Faithfulness}
    \caption{Prompt used to evaluate the \textit{Faithfulness} dimension.}
    \label{fig:prompt_faithfulness}
\end{figure*}

\begin{figure*}[!htb]
    \input{Tables/prompts/Main_Issues}
    \caption{Prompt used to evaluate the \textit{Main Issues} dimension.} 
    \label{fig:prompt_main_issues}
\end{figure*}

\begin{figure*}[!htb]
    \input{Tables/prompts/Sub_Issues}
    \caption{Prompt used to evaluate the \textit{Sub-issues} dimension.}
    \label{fig:prompt_sub_issues}
\end{figure*}

\begin{figure*}[!htb]
    \input{Tables/prompts/Resolution}
    \caption{Prompt used to evaluate the \textit{Resolution} dimension.}
    \label{fig:prompt_resolution}
\end{figure*}

\section{Human Evaluation}
\label{sec:human_evaluation_details}

\subsection{Full results}
The full human evaluation results for the DECODA dataset are detailed in Table \ref{tab:human_eval_results_full}, categorized into three groups: all 20 dialogues, the 10 shortest dialogues and the 10 longest dialogues.

\begin{table*}[!htb]
\footnotesize
\centering
  \begin{tabular}{lccccc}
\toprule
\textbf{} & \textbf{Faithfulness} & \textbf{Main Issues} & \textbf{Sub-issues} & \textbf{Resolution} & \textbf{Overall} \\ 
\midrule
\rowcolor{Gray}
\textbf{All Dialogues} & & & & & \\
Reference   & 3.72$\pm$1.12 & 4.15$\pm$1.07 & 3.22$\pm$1.86 & 3.25$\pm$1.60 & 3.68$\pm$1.36 \\ 
BARThez     & 2.17$\pm$1.43 & 2.32$\pm$1.57 & 1.00$\pm$0.0 & 2.22$\pm$1.54 & 2.17$\pm$1.49 \\ 
3.5-WL      & 4.02$\pm$1.17 & 4.30$\pm$1.18 & 2.67$\pm$1.73 & 4.18$\pm$1.28 & 4.10$\pm$1.28 \\ 
3.5-HGR$\rightarrow$ WL  & 3.97$\pm$1.15 & \textbf{4.47$\pm$0.89} & \textbf{3.67$\pm$1.73} & \textbf{4.20$\pm$1.13} & \textbf{4.19$\pm$1.12} \\ 
4-WL        & \textbf{4.13$\pm$0.96} & 4.37$\pm$0.80 & 1.44$\pm$1.33 & 2.70$\pm$1.79 & 3.62$\pm$1.53 \\ 
4-HGR$\rightarrow$ WL    & 3.93$\pm$1.02 & 4.30$\pm$0.96 & 2.67$\pm$1.66 & 4.05$\pm$1.27 & 4.03$\pm$1.16 \\ 
\midrule
\rowcolor{Gray}
\textbf{10 Short Dialogues} & & & & & \\
Reference   & 3.53$\pm$1.07 & 4.07$\pm$1.11 & 5.00 & 3.17$\pm$1.70 & 3.60$\pm$1.37 \\ 
BARThez     & 2.27$\pm$1.39 & 2.20$\pm$1.63 & 1.00 & 2.27$\pm$1.57 & 2.23$\pm$1.51 \\ 
3.5-WL      & 3.83$\pm$1.32 & 4.13$\pm$1.33 & 1.00 & \textbf{4.47$\pm$1.04} & 4.11$\pm$1.29 \\ 
3.5-HGR$\rightarrow$ WL  & 4.27$\pm$1.01 & \textbf{4.67$\pm$0.80} & 5.00 & 4.37$\pm$1.10 & \textbf{4.44$\pm$0.98} \\ 
4-WL        & \textbf{4.37$\pm$0.89} & 4.47$\pm$0.78 & 5.00 & 4.17$\pm$1.21 & 4.34$\pm$0.97 \\ 
4-HGR$\rightarrow$ WL    & 4.10$\pm$0.92 & 4.63$\pm$0.61 & 5.00 & 4.20$\pm$1.30 & 4.32$\pm$1.00 \\ 
\midrule
\rowcolor{Gray}
\textbf{10 Longest Dialogues} & & & & & \\
Reference   & 3.90$\pm$1.16 & 4.23$\pm$1.04 & 3.00$\pm$1.85 & 3.33$\pm$1.52 & 3.76$\pm$1.36 \\ 
BARThez     & 2.07$\pm$1.48 & 2.43$\pm$1.52 & 1.00$\pm$0.0 & 2.17$\pm$1.53 & 2.12$\pm$1.48 \\ 
3.5-WL      & \textbf{4.20$\pm$1.00} & \textbf{4.47$\pm$1.01} & 2.88$\pm$1.73 & 3.90$\pm$1.45 & \textbf{4.08$\pm$1.27} \\ 
3.5-HGR$\rightarrow$ WL  & 3.67$\pm$1.21 & 4.27$\pm$0.94 & \textbf{3.50$\pm$1.77} & \textbf{4.03$\pm$1.16} & 3.95$\pm$1.19 \\ 
4-WL        & 3.90$\pm$0.99 & 4.27$\pm$0.83 & 1.00$\pm$0.0 & 1.23$\pm$0.77 & 2.96$\pm$1.65 \\ 
4-HGR$\rightarrow$ WL    & 3.77$\pm$1.10 & 3.97$\pm$1.13 & 2.38$\pm$1.51 & 3.90$\pm$1.24 & 3.76$\pm$1.24 \\ 
\bottomrule
\end{tabular}
  \caption{
Mean ($\pm$ std) scores for \textit{Faithfulness}, \textit{Main issues}, \textit{Sub-issues}, \textit{Resolution}, and \textit{Overall} of each summary assessed by three native human evaluators on the 10 shortest and 10 longest dialogues in the DECODA dataset. The best results for each criterion are shown in bold.}\label{tab:human_eval_results_full}
\end{table*}

\subsection{Radar chart}

\begin{figure}[!htb]
\begin{center}
\includegraphics[width=.65\columnwidth]{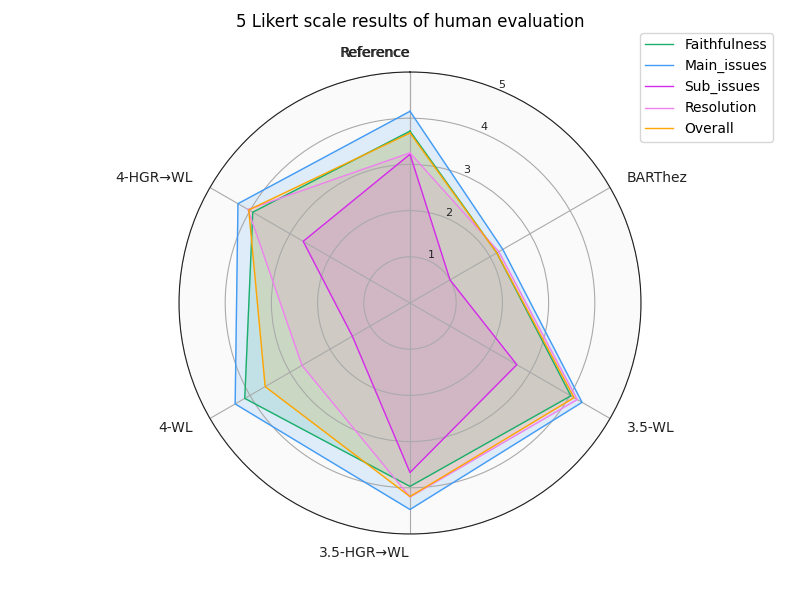}
\caption{DECODA human evaluation results on a 5-point Likert scale. The radar chart compares six summaries (including reference summaries and those generated by various systems) across four criteria and the overall score.}
\label{fig:radar_chart}
\end{center}
\end{figure}

The results for the 20 selected samples are presented in the radar chart in Figure \ref{fig:radar_chart}. This visualization shows the evaluation of six summaries, including both reference summaries and those generated by different systems, across four criteria and the overall scores. The visual representation indicates that GPT-generated summaries are generally preferred by human annotators over those from the task-specific pre-trained model BARThez, and that they outperform the reference summaries on some criteria, such as \textit{Faithfulness}.

\section{Example Analysis of Outputs}
\label{sec:appendix_examples}

\paragraph{DialogSum}

We present three examples in the following tables. We observe that, when using prompts designated as \texttt{HG(R)}, GPT models tend to generate long summaries that are too precise and detailed, sometimes exceeding the length of the original conversation. Yet the instructions explicitly state that the summary should not exceed 20\% of the original dialogue. 
Consequently, GPT models fail to fully adhere to prescribed human summarization guidelines.

In detail, Table~\ref{tab:outputs_analysis} reveals that the summary generated by \chatgpt{} with the \texttt{WordLimit} prompt is of higher quality than that produced by \gptfour{}. It talks not only about the benefits of the job, but also about its demanding schedule. However, for \chatgpt{}, we found \texttt{HG$\rightarrow$WL} to be quite good, whereas \texttt{HGR$\rightarrow$WL} seems to tell the story from \textit{Person1}'s point of view instead of \textit{Person2}'s. As for \gptfour{}, the summary predicted by \texttt{HGR$\rightarrow$WL} is not very good, as it only talks about the benefits, omitting the demanding schedule of this new job.

In Table~\ref{tab:outputs_analysis_2}, the \texttt{WordLimit} prompt yields well-sized summaries for both GPT models, whereas other prompts produce excessively long outputs -- particularly notable given the short length of the source conversation.
In Table~\ref{tab:outputs_analysis_3}, the predictions of \texttt{WordLimit} prompt are good for both GPT models, as does the \texttt{HG(R)$\rightarrow$WL} prompt.

\begin{table*}[!htb]
    \footnotesize
    \centering
    \resizebox{\textwidth}{!}{%
    \begin{tabular}{|p{16cm}|}
\hline
\multicolumn{1}{|c|}{\textbf{test\_20}} \\
\hline
\textbf{Dialogue:} \\
\hangindent=1em
\#Person1\#: Hey, Frank. I heard you got a new job.
\#Person2\#: Yeah, Judy. I will be working for the Post Office. It's not a bad job.
\#Person1\#: Is it true that you have a heavy work schedule?
\#Person2\#: That's right. I am supposed to work at 5am everyday, and I only get 45 minutes for lunch.
\#Person1\#: So, why did you apply for such a demanding job?
\#Person2\#: Well, the government offers its employees excellent health insurance benefits.
\#Person1\#: Oh, I see. And can your family members utilize the health insurance, too?\#Person2\#: Yeah, that's the best part. All of my children can get free medical care.
\#Person1\#: That's a great employment benefit!
\#Person2\#: Now you know why I wanted to work for the Post Office! \\
[1ex]

\textbf{Reference Summaries:} \\
\textbf{Summary1:} 
Frank got a new job and is telling Judy not only the heavy schedule but also the benefits of this job. \\
\textbf{Summary2:} Frank explains to Judy the reason why he took a job at the Post office is that the government offers excellent health insurance benefits for its employees. \\
\textbf{Summary3:} Frank describes his new job's schedule and insurance at the Post Office to Judy. \\
\hline

\textbf{\chatgpt{} Generated Summaries} \\[0.5ex]
\hline
\textbf{3.5-WL:}  Person2 got a new job at the Post Office with a demanding work schedule, but the excellent health insurance benefits for family members made it worth it.\\
\hdashline
\textbf{3.5-HG:} Person1 informs Person2 that they heard about Person2's new job. Person2 confirms that they will be working for the Post Office and mentions that it is not a bad job. Person1 asks if it is true that Person2 has a heavy work schedule. Person2 confirms and explains that they are supposed to work at 5am every day and only have 45 minutes for lunch. Person1 asks why Person2 applied for such a demanding job. Person2 explains that the government offers excellent health insurance benefits to its employees. Person1 understands and asks if Person2's family members can also utilize the health insurance. Person2 confirms that all of their children can receive free medical care. Person1 acknowledges that it is a great employment benefit. Person2 concludes by stating that this is why they wanted to work for the Post Office.
 \\
\hdashline
\textbf{3.5-HGR:} Person1 informs Person2 that they heard about Person2's new job. Person2 confirms and mentions that they will be working for the Post Office. Person1 asks about the work schedule, and Person2 explains that they have to work at 5am every day and have a short lunch break. Person1 questions why Person2 applied for such a demanding job, and Person2 explains that the government offers excellent health insurance benefits. Person1 realizes that family members can also utilize the health insurance, and Person2 confirms. Person1 comments on the great employment benefit, and Person2 explains that this is why they wanted to work for the Post Office.
 \\
\hline
\textbf{3.5-HG$\rightarrow$WL:} Person2 tells Person1 about their new job at the Post Office, mentioning the demanding work schedule but highlighting the excellent health insurance benefits for their family.
 \\
\hdashline
\textbf{3.5-HGR$\rightarrow$WL:} Person1 learns about Person2's new job at the Post Office, which has demanding hours but offers excellent health insurance benefits for family members.
 \\
\hline

\textbf{\gptfour{} Generated Summaries} \\[0.5ex]
\hline
\textbf{4-WL:} Frank got a new job at the Post Office for its excellent health insurance benefits for his family.
 \\
\hdashline
\textbf{4-HGR:} In the conversation, Judy learns that Frank has secured a new job at the Post Office, which demands an early start at 5am and allows only a short lunch break. Despite the rigorous schedule, Frank's decision to take up the job is driven by the excellent health insurance benefits offered by the government to its employees. This benefit extends to his children, providing them with free medical care, which Judy acknowledges as a significant employment advantage.
 \\
\hline
\textbf{4-HGR$\rightarrow$WL:} Frank accepts a demanding Post Office job for its excellent health insurance benefits, covering his children's medical care.
 \\
\hline
\end{tabular}
    }
    \caption{Input dialogue, reference summaries, and model-generated summaries for test sample \textit{test\_20}.} 
    \label{tab:outputs_analysis}
\end{table*}

\begin{table*}[!htb]
    \small
    \centering
    \resizebox{\textwidth}{!}{%
    \begin{tabular}{|p{16cm}|}
\hline
\multicolumn{1}{|c|}{\textbf{test\_151}} \\
\hline
\textbf{Dialogue:} \\
\hangindent=1em
\#Person1\#: Have you enjoyed your weekend?
\#Person2\#: Yes, it's been marvelous. It really was very kind of you to invite me. I hope it hasn't been too much trouble.
\#Person1\#: Not at all. We've really enjoyed having you. I hope you'll come and stay again next time you're in Changsha. \\
[1ex]

\textbf{Reference Summaries:} \\

\textbf{Summary 1:} \#Person2\# tells \#Person1\# \#Person2\# enjoyed the weekend organized by \#Person1\#. \\
\textbf{Summary 2:} \#Person1\# and \#Person2\# has enjoyed a nice weekend together. \\
\textbf{Summary 3:} \#Person1\# invited \#Person2\# to spend the weekend and \#Person2\# has enjoyed it. \\
\hline

\textbf{\chatgpt{} Generated Summaries} \\[0.5ex]
\hline
\textbf{3.5-WL:} Person2 had a marvelous weekend and thanked Person1 for the invitation, hoping to stay again in the future. \\
\hdashline
\textbf{3.5-HG:} Person1 asks Person2 if they have enjoyed their weekend. Person2 responds positively, saying it has been marvelous and thanking Person1 for the invitation. Person2 expresses concern that it may have been too much trouble, but Person1 reassures them that it was not and that they have enjoyed having Person2. Person1 also invites Person2 to come and stay again next time they are in Changsha.
 \\
\hdashline
\textbf{3.5-HGR:} Person1 asks Person2 if they have enjoyed their weekend. Person2 responds positively, saying it has been marvelous and thanking Person1 for the invitation. Person1 reassures Person2 that it was not trouble at all and expresses their enjoyment of having Person2. Person1 also invites Person2 to stay again in Changsha in the future.
 \\
\hline
\textbf{3.5-HG$\rightarrow$WL:} Person2 had a marvelous weekend and thanked Person1 for the invitation. Person1 reassured them and invited them to visit again.
 \\
\hdashline
\textbf{3.5-HGR$\rightarrow$WL:} Person2 had a marvelous weekend and thanked Person1 for the invitation. Person1 enjoyed having Person2 and invited them to visit again in Changsha.
 \\
\hline

\textbf{\gptfour{} Generated Summaries} \\[0.5ex]
\hline
\textbf{4-WL:} Person1 invites Person2 for a weekend stay in Changsha, which Person2 thoroughly enjoys.
 \\
\hdashline
\textbf{4-HGR:} Person1 inquires if Person2 enjoyed their weekend, to which Person2 responds affirmatively, expressing gratitude for the invitation and concern about any potential inconvenience caused. Person1 dismisses any notion of trouble and extends another invitation for Person2 to visit again when they are next in Changsha.
 \\
\hline
\textbf{4-HGR$\rightarrow$WL:} Person1 invites Person2 again after a pleasant weekend visit in Changsha.
 \\
\hline
\end{tabular}
    }
    \caption{Input dialogue, reference summaries, and model-generated summaries for test sample \textit{test\_151}.}
    \label{tab:outputs_analysis_2}
\end{table*}

\begin{table*}[!htb]
    \footnotesize
    \centering
            \resizebox{\textwidth}{!}{%
    \begin{tabular}{|p{16cm}|}
\hline
\multicolumn{1}{|c|}{\textbf{test\_440}} \\
\hline
\textbf{Dialogue:} \\
\hangindent=1em
\#Person1\#: Man, I'm freaking out! You gotta help me! 
\#Person2\#: Whoa, whoa, take it easy, relax. Geez, you're sweating like a pig! What's going on? 
\#Person1\#: I can't go through with this! I just can't! I'm not ready for marriage! What was I thinking? I'm only thirty five years old! I've got my entire life ahead of me, adventures waiting! I can't settle down yet! 
\#Person2\#: What are you talking about? It wasn't more than a month ago that you were rambling on about how you are tired of living the life of a bachelor and how you envy your friends that have a family! 
\#Person1\#: I know. I know! 
\#Person2\#: Let's think this through. First of all, you cannot leave Amy at the altar. Not only will she be humiliated and kill you, but she is the woman that you love and the woman of your dreams! Why would you want to end that or jeopardize it like this? Second of all, you are just getting cold feet. You know deep down inside that you want to marry her, so cut the crap and do it! 
\#Person1\#: You're right. I'm being crazy! Ok, I'm ready, let's do this! 
\#Person2\#: Great! Phew! That was a close one! You had me going there for a minute I thought I was gonna have to slap some sense into you. \\
[1ex]

\textbf{Reference Summaries:} \\
\textbf{Summary 1:} \#Person1\# is not ready for marriage. \#Person2\# reminds him of what he said a month ago and what will happen if he leaves Amy. After listening to \#Person2\#, \#Person1\# is ready for marriage. \\
\textbf{Summary 2:} \#Person1\# feels anxious about future marriage and sweats a lot. \#Person2\# helps \#Person1\# to think through the current situation and get ready for the marriage. \\
\textbf{Summary 3:} \#Person1\# is freaking out before his marriage with Amy and \#Person2\# helps him get over it by helping him think through all the details. \\
\hline

\textbf{\chatgpt{} Generated Summaries} \\[0.5ex]
\hline
\textbf{3.5-WL:} Person 1 is having second thoughts about getting married, but Person 2 convinces them to go through with it. \\
\hdashline
\textbf{3.5-HG:} Person1 is feeling anxious and unsure about getting married. They express their concerns about settling down at their age and missing out on adventures. Person2 tries to calm them down and reminds them of their previous desire to have a family. Person2 also emphasizes the importance of not leaving Amy at the altar and reminds Person1 of their love for her. Person1 realizes they are just getting cold feet and agrees to go through with the wedding. Person2 is relieved and jokes about having to slap some sense into Person1.
 \\
\hdashline
\textbf{3.5-HGR:} Person1 is panicking and expressing doubts about getting married. Person2 tries to calm Person1 down and reminds them of their previous desire to settle down. Person2 also emphasizes the importance of not leaving Amy at the altar and encourages Person1 to go through with the wedding. Person1 eventually agrees and decides to proceed with the marriage. Person2 expresses relief and jokes about having to intervene if Person1 didn't change their mind.
 \\
\hline
\textbf{3.5-HG$\rightarrow$WL:} Person1 is anxious about getting married but Person2 reassures them and reminds them of their love for their partner.
 \\
\hdashline
\textbf{3.5-HGR$\rightarrow$WL:} Person1 has doubts about getting married, but Person2 convinces them to go through with it, emphasizing the importance of not leaving Amy at the altar.
 \\
\hline

\textbf{\gptfour{} Generated Summaries} \\[0.5ex]
\hline
\textbf{4-WL:} A man panics before his wedding, but his friend calms him down and convinces him to go through with it.
 \\
\hdashline
\textbf{4-HGR:} In a state of panic, Person1 expresses his fear of getting married to Person2, despite being thirty-five years old. He feels he's not ready to settle down and still has many adventures to experience. Person2 reminds him of his previous desire to leave bachelorhood and his envy of his friends with families. Person2 also emphasizes that Person1 cannot leave his fiancée, Amy, at the altar, as it would not only humiliate her but also jeopardize their relationship. Person2 convinces Person1 that his fear is just a case of cold feet and that he truly wants to marry Amy. Accepting this, Person1 calms down and decides to proceed with the wedding, much to Person2's relief.
 \\
\hline
\textbf{4-HGR$\rightarrow$WL:} Person1, fearing marriage, is reassured by Person2 and decides to proceed with his wedding to Amy.
 \\
\hline
\end{tabular}
    }
    \caption{Input dialogue, reference summaries, and model-generated summaries for test sample \textit{test\_440}.} 
    \label{tab:outputs_analysis_3}
\end{table*}

\paragraph{DECODA}

We present an example of the sample \textit{FR\_20091112\_RATP\_SCD\_0826} in Table \ref{tab:decoda_outputs_analysis}. We note that the synopses generated with \texttt{WordLimit} and \texttt{HG(R)$\rightarrow$WL} prompts for both GPT models are good. The synopses generated with \texttt{HG(R)} prompt follow the structure of the given instructions: first talk about the main issue, then the sub-issues of the conversation, and finally the resolution of the call, even the predictions are long, but make the two-step prompt approach -- \texttt{HG(R)$\rightarrow$WL} generating quite good synopses.

\begin{table*}[!htb]
    \footnotesize
    \centering
            \resizebox{\textwidth}{!}{%
    \begin{tabular}{|p{15.5cm}|}
\hline
\textbf{Dialogue:}    \\
\hangindent=1em
      $<$Spk\_A$>$ bonjour \\
      $<$Spk\_B$>$ oui bonjour madame \\
      $<$Spk\_B$>$ je vous appelle pour avoir des horaires de train en+fait c' est pas pour le métro je sais pas si vous pouvez me les donner ou pas  \\
      $<$Spk\_A$>$ trains SNCF  \\
      $<$Spk\_B$>$ oui trains SNCF oui  \\
      $<$Spk\_A$>$ vous prenez quelle ligne monsieur  \\
      $<$Spk\_B$>$ euh la ligne euh enfin en+fait c' est pas SNCF enfin c' est Île-de-France quoi je sais pas comment ils appellent ça  \\
      $<$Spk\_B$>$ RER voilà c' est pour les RER  \\
      $<$Spk\_B$>$ voilà et euh je prends la ligne euh Meaux de Meaux pour aller à Paris je sais plus c' est laquelle c' est la  \\
      $<$Spk\_B$>$ E je crois  \\
      $<$Spk\_A$>$ d'accord et vous voulez donc les horaires euh pour quel jour et à quel moment monsieur \\ $<$Spk\_B$>$ euh là pour euh tout à l' heure euh pour euh aux environs de dix-sept heures en partant de la gare de Meaux \\
      $<$Spk\_B$>$ euh vers la Gare+de+l'Est à Paris \\
      $<$Spk\_A$>$ alors vous partez de Meaux et vous allez donc à la Gare+de+l'Est $<$Spk\_A$>$ et vous voudriez les horaires $<$Spk\_B$>$ voilà \\
      $<$Spk\_A$>$ vers euh dix-sept heures $<$Spk\_A$>$ alors moi je peux regarder ce+que $<$Spk\_B$>$ ouais dix-sept heures ouais \\
      $<$Spk\_A$>$ j' ai comme horaires un instant monsieur $<$Spk\_A$>$ s'il+vous+plaît $<$Spk\_B$>$ d'accord il y a \\
      $<$Spk\_B$>$ pas de souci \\
      $<$Spk\_A$>$ monsieur s'il+vous+plaît \\
      $<$Spk\_B$>$ oui \\
      $<$Spk\_A$>$ donc voilà ce+que j' ai comme horaires moi vous avez donc un départ à la gare de Meaux donc à seize heures quarante-neuf \\
      $<$Spk\_B$>$ seize heures quarante-neuf d'accord \\
      $<$Spk\_A$>$ et après vous avez celui de dix-sept heures dix-neuf \\
      $<$Spk\_B$>$ alors seize heures quarante-neuf dix-sept heures dix-neuf $<$Spk\_A$>$ OK d'accord $<$Spk\_B$>$ oui \\
      $<$Spk\_B$>$ ben je vous remercie \\
      $<$Spk\_A$>$ mais je vous en prie $<$Spk\_B$>$ bonne journée madame \\
      $<$Spk\_A$>$ merci à vous aussi monsieur au+revoir \\
      $<$Spk\_B$>$ au+revoir \\
[1ex]

\textbf{Reference Synopses:} \\
\textbf{synopsis 1:} Demande d'horaires de train de la gare de Maux à la gare de l'Est à une heure donnée. \\
\textbf{synopsis 2:} horaires RER E de Meaux à la Gare de l'Est \\
\hline

\textbf{\chatgpt{} Generated Summaries} \\[0.5ex]
\hline
\textbf{3.5-WL:} Un homme appelle pour demander les horaires des trains RER entre Meaux et Paris. Il obtient les horaires pour un départ à 16h49 et à 17h19. \\
\hdashline
\textbf{3.5-HG:} Le client appelle pour obtenir des horaires de train pour la ligne RER de Meaux à Paris. Il demande les horaires pour un départ aux environs de 17 heures. L'agent recherche les horaires et lui donne deux options : un départ à 16h49 et un autre à 17h19. Le client remercie l'agent et l'appel se termine. \\
\hdashline
\textbf{3.5-HGR:} Le client appelle pour obtenir des horaires de train pour la ligne RER de Meaux à Paris. Il demande les horaires pour un départ aux environs de 17 heures. L'agent recherche les horaires et lui donne deux options : un départ à 16h49 et un autre à 17h19. Le client remercie l'agent et l'appel se termine. \\

Le problème principal de l'appel est la demande des horaires de train pour la ligne RER de Meaux à Paris. Les sous-problèmes incluent la clarification de la ligne de train (RER) et la spécification de l'heure de départ (aux environs de 17 heures). \\

L'appel est résolu au cours de cet appel, car l'agent fournit les horaires demandés au client. \\
\hline
\textbf{3.5-HG$\rightarrow$WL:} Le client appelle pour obtenir les horaires de train de la ligne RER de Meaux à Paris. L'agent lui propose deux options de départ : 16h49 ou 17h19. L'appel est résolu. \\
\hdashline
\textbf{3.5-HGR$\rightarrow$WL:} Le client appelle pour obtenir les horaires de train de la ligne RER de Meaux à Paris. L'agent lui propose deux options de départ : 16h49 ou 17h19. L'appel se termine avec satisfaction. \\
\hline

\textbf{\gptfour{} Generated Summaries} \\[0.5ex]
\hline
\textbf{4-WL:} Un homme appelle pour connaître les horaires de train de la ligne RER E entre Meaux et Paris pour le jour même. \\
\hdashline
\textbf{4-HGR:} Le client a appelé pour obtenir des informations sur les horaires des trains RER de la ligne E, de Meaux à Paris Gare de l'Est, pour le jour même aux environs de 17 heures. L'agent a pu fournir les horaires demandés, indiquant un départ à 16h49 et un autre à 17h19. Le client a remercié l'agent pour les informations fournies et a terminé l'appel. Le problème principal de l'appel, à savoir la demande d'horaires de train, a été résolu lors de cet appel. \\
\hline
\textbf{4-HGR$\rightarrow$WL:} Un client a demandé les horaires des trains RER E de Meaux à Paris, l'agent a fourni les informations nécessaires. \\
\hline
\end{tabular}
    }
    \caption{Input dialogue, reference synopses, and model-generated summaries for DECODA test sample \textit{FR\_20091112\_RATP\_SCD\_0826}.}
    \label{tab:decoda_outputs_analysis}
\end{table*}

\section{Output Length}
\label{sec:appendix_output_length}

\begin{figure*}[!htb]
\centering
\resizebox{\textwidth}{!}{%
\includegraphics[width=\textwidth]{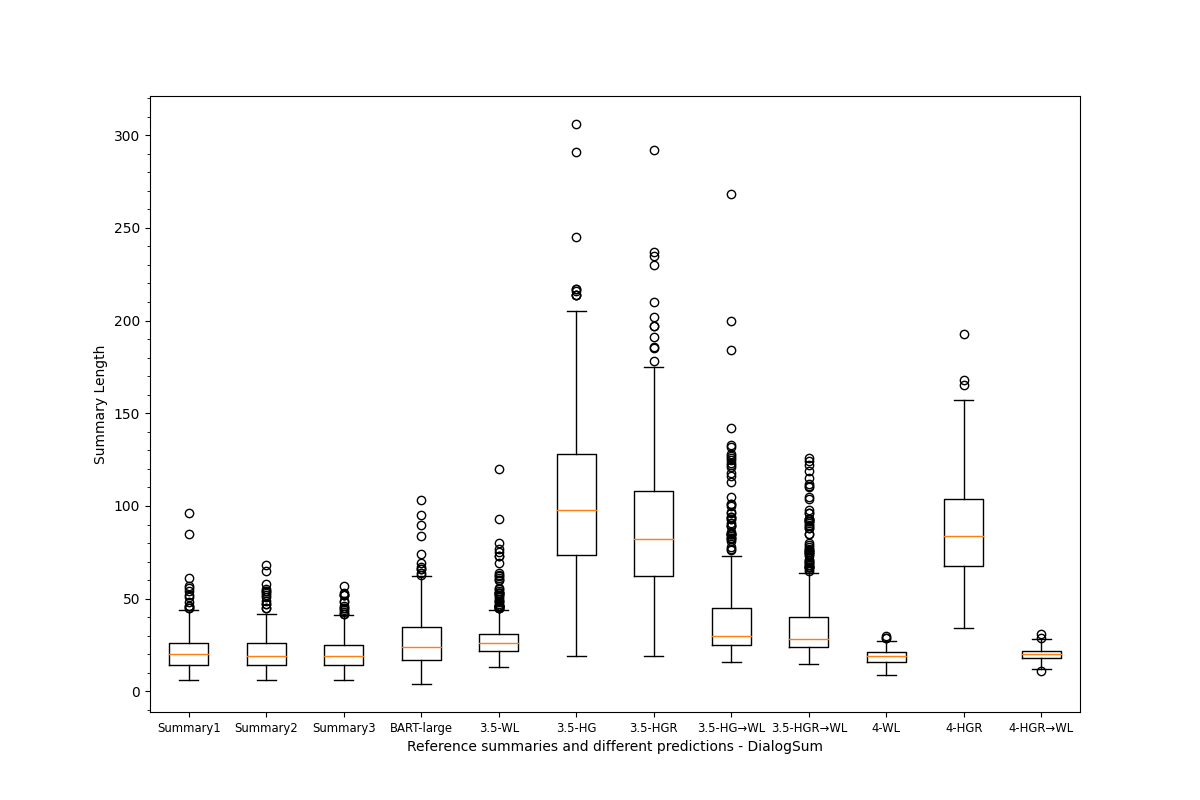}
}
\caption{Distribution of generated summary lengths on the DialogSum test set, compared to the three reference summaries.}
\label{fig:summ_length_openai_dialogsum}
\end{figure*}

\begin{figure*}[!htb]
\begin{center}
\includegraphics[width=\textwidth]{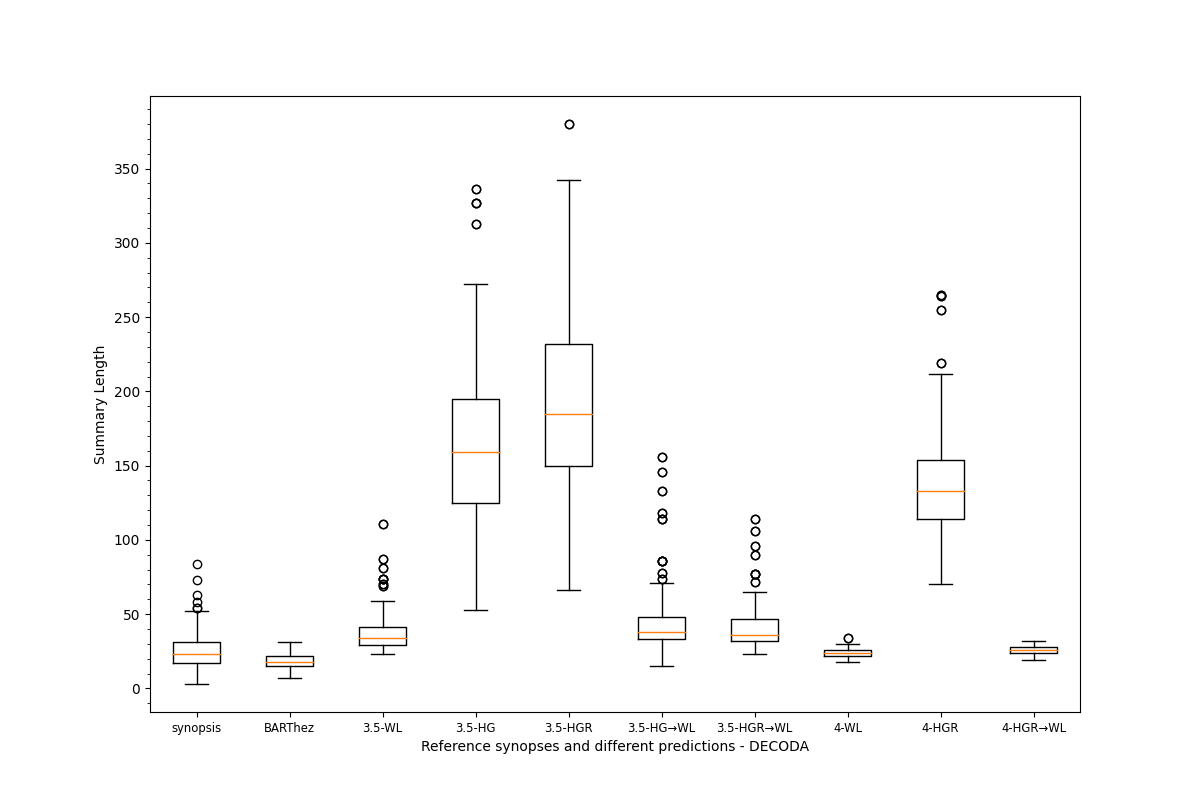}
\caption{Distribution of generated summary lengths on the DECODA test set, compared to the reference synopses.}
\label{fig:summ_length_openai_decoda}
\end{center}
\end{figure*}

Figure~\ref{fig:summ_length_openai_dialogsum} and Figure~\ref{fig:summ_length_openai_decoda} present summary length statistics for different prompts on the DialogSum and DECODA datasets, respectively, compared to reference summaries.

The summaries generated by \chatgpt{} are consistently longer than the reference summaries. Applying the \texttt{WordLimit} prompt reduced the average output length, and using it as a second step after the \texttt{HG(R)} prompt produced more concise summaries (\texttt{HGR$\rightarrow$WL}).

Regarding \gptfour{}, it demonstrated better adherence to length instructions. The distributions for \texttt{4-WL} and \texttt{4-HGR$\rightarrow$WL} are more constrained in both figures, while \texttt{4-HGR} exceeds the reference summary length.

\section{Manual Error Analysis}

\subsection{DialogSum Discrepancy Data Points}

Table \ref{tab:discrepancy_dialogsum_4_HGR_WL} presents DialogSum cases with the most significant scoring discrepancies (low ROUGE-L but high BERTScore) for \texttt{4-HGR$\rightarrow$WL} summaries.

\begin{table*}[!htb]
    \footnotesize
            \resizebox{\textwidth}{!}{%
    \begin{tabular}{l>{\raggedright}p{7.4cm}>{\raggedright}p{3.5cm}llll}
\toprule
\textbf{ID}  & \textbf{References}    & \textbf{Prediction}    & \textbf{RL}  & \textbf{BS} & \textbf{Diff.} \\
\midrule
246 & ["Tim"s project is about how to lead a greener life. \#Person1\#   consults about his motivation and his findings at school. Tim finds there are   many aspects that can be improved for his school.", "Tims is   talking about his greener life. He now rides his bike to school and   recommends teachers to send electrical paper rather than paper ones. And he   thinks the school shouldn"t have machines with bottled drinks.",   "Tim chooses a green-life-related project because he wants to see what he   could do in his own life. He"s started being greener with himself and   influences others."]   & \#Person2\# proposes eco-friendly changes at school and personal life,   inspired by environmental science lessons.                                & 6.77         & 86.81             & 80.04    \\
\midrule
232 & ["\#Person1\# asks \#Person2\# about past experiences of living in different   cities around the world.", "\#Person1\# is asking \#Person2\# about the places   that \#Person2\# has stayed or lived.", "\#Person1\# and \#Person2\# talk   about \#Person2\#"s experience. \#Person2\# moves a lot. \#Person1\# finds it   interesting."]   & \#Person2\# shares their life journey from Chicago to Los Angeles and   Munich with \#Person1\#.       & 9.29         & 89.26             & 79.96    \\
\midrule
283 & ["\#Person2\# wants to buy some clothes but \#Person1\#"s recommendation   doesn"t interest \#Person2\#.", "\#Person1\# gives \#Person2\# a few   suggestions on clothes but none is taken.", "\#Person2\# is not happy with   \#Person1\#"s recommendeds."]    & Salesperson suggests brown, silk, and cotton dresses to a customer who   prefers a white dress for the warm season.                                & 4.17         & 83.96             & 79.79     \\
\midrule
433 & ["Amy is talking about her first job experience and later job with   \#Person1\#.", "Amy tells \#Person1\# about her first job experience and her   current job.", "Amy shares her former work experience and her current job   with \#Person1\#."]      & Amy, after starting as a secretary in 1998, became a Personnel Manager,   then moved to Sales due to salary dissatisfaction.                       & 6.19         & 85.7              & 79.51    \\
\midrule
379 & ["\#Person1\# interviews Vet, a high school girl who has become a mother.   She shows her love to her daughter and talks about her future plans and gives   advice to teens.", "Vet tells her own experience as a young mother and how   her daughter changes her. Then she tells \#Person1\# her future goals and   dreams. And she also gives teens advice.", "\#Person1\# is interviewing Vet and   asking her about the ways she deals with stress as a young mother, feelings   of having a baby, plans for the future, her favorite quote, and advice for   teens."]      & \#Person2\# discusses stress coping mechanisms, the impact of motherhood,   future ambitions, and advises teenagers on self-love and independence. & 11.83         & 89.75             & 77.92     \\
\bottomrule
\end{tabular}
    }
    \caption{\label{tab:discrepancy_dialogsum_4_HGR_WL} Data points where GPT-generated summaries (\texttt{4-HGR$\rightarrow$WL}) show the greatest discrepancy in the DialogSum corpus, with low ROUGE-L but high BERTScore values. RL and BS are the average scores of three references.} 
\end{table*}

\subsection{DECODA Discrepancy Data Points}

Table \ref{tab:discrepancy_decoda_4_HGR_WL_translation_en} displays English translations of DECODA cases showing the most significant scoring discrepancies (low ROUGE-L but high BERTScore) for \texttt{4-HGR$\rightarrow$WL} summaries.

\begin{table*}[htb]
\footnotesize
        \resizebox{\textwidth}{!}{%
    \begin{tabular}{lp{5.4cm}p{5.4cm}llll}
\toprule
\textbf{ID}  & \textbf{Reference}    & \textbf{Prediction}    & \textbf{RL}  & \textbf{BS} & \textbf{Diff.}  \\
\midrule
10 & Inquiry about a corporate account order. Transfer to the relevant department. & A representative from the Ministry of Foreign Affairs is organizing a seminar in Paris and contacts RATP to obtain transport cards for participants. & 0.00 & 70.14 & 70.14 \\
\midrule
135 & Route request from Drancy station to Gare du Nord. Clarification of explanations following a previous call. & A customer confused about RER B connection to reach Drancy is reassured they can travel directly from Bourg-la-Reine. & 4.55 & 70.29 & 65.75 \\
\midrule
29 & Inquiry about purchasing tickets for school groups. Transfer to the relevant department. & A school association representative calls to understand public transport procedures for students. The agent proposes a solution and transfers the call. & 8.51 & 72.83 & 64.32 \\
\midrule
68 & Wrong number - caller meant to dial social security at 36 46 but misdialed & A customer mistakes RATP customer service for Social Security regarding a health card issue in the Antilles. & 4.65 & 66.03 & 61.38 \\
\midrule
140 & No ImaginR card subscription by mail, only at ticket counters in all stations & A customer requested an Imagine+R application by mail. The agent explained it's impossible but provided the service number for more information. & 4.17 & 65.54 & 61.38 \\
\bottomrule
\end{tabular}

    }
\caption{\label{tab:discrepancy_decoda_4_HGR_WL_translation_en} Data points where GPT-generated summaries (\texttt{4-HGR$\rightarrow$WL}) show the greatest discrepancy in the DECODA corpus, with low ROUGE-L but high BERTScore values. This is the translated version; the original texts are in Table \ref{tab:discrepancy_decoda_4_HGR_WL} in the Appendix.}
\end{table*}

Tables \ref{tab:discrepancy_decoda_4_WL} and \ref{tab:discrepancy_decoda_4_HGR_WL} present original French DECODA cases with the most significant scoring discrepancies (low ROUGE-L but high BERTScore) for \texttt{4-WL} and \texttt{4-HGR$\rightarrow$WL} summaries, respectively, as analyzed in Section \ref{subsec:manual_error_analysis}.

\begin{table*}[!htb]
\footnotesize
        \resizebox{\textwidth}{!}{%
\begin{tabular}{lp{5.4cm}p{5.4cm}llll}
\toprule
\textbf{ID}  & \textbf{Reference}    & \textbf{Prediction}    & \textbf{RL}  & \textbf{BS} & \textbf{Diff.}  \\
\midrule
10  & Demande d'information sur une commande grands comptes. Transfert vers le   service concerné.  & Un organisateur de séminaire cherche à  obtenir des cartes de   transport pour ses invités internationaux auprès de la RATP.       & 0      & 71.13    & 71.13      \\
\midrule
110 & Demande de renseignements suite à  la procédure suivi après la   réception d'un courrier d'amende. Transfert vers le service concerné.  & Une cliente appelle le service clientèle pour comprendre pourquoi elle   doit payer cinq euros suite à  une amende de bus contestée et   justifiée. & 8.33 & 70.86    & 62.53      \\
\midrule
107 & demande procédure pour remboursement suite à  achat tickets car   envoi tardif du passe ImaginR, attente au-delà  de 3 semaines donc   possible, appeler ImaginR ou par mail  & Une cliente appelle pour demander le remboursement d'une carte Orange   achetée en attendant sa carte Imagine+R, arrivée en retard.    & 8.00   & 69.35    & 61.35      \\
\midrule
32  & circulation du T2  & Un utilisateur se renseigne sur la mise en service du T+deux de la Porte  de Versailles à  Val d'Issy.    & 8.70  & 69.85    & 61.15   \\
\midrule
29  & Demande de renseignements sur l'achat de ticket pour des groupes   scolaires. Transfert vers le service concerné.  & Un représentant d'une association départementale cherche à    organiser le transport de plus de 100 classes via les transports en commun.     & 10.26 & 71.05    & 60.80    \\
\bottomrule
\end{tabular}
}
\caption{\label{tab:discrepancy_decoda_4_WL} Data points where GPT-generated summaries (\texttt{4-WL}) show the greatest discrepancy in the DECODA corpus, with low ROUGE-L but high BERTScore values.}
\end{table*}

\begin{table*}[!htb]
\footnotesize
\begin{tabular}{lp{5.4cm}p{5.4cm}llll}
\toprule
\textbf{ID}  & \textbf{Reference}    & \textbf{Prediction}    & \textbf{RL}  & \textbf{BS} & \textbf{Diff.}  \\
\midrule
10 & Demande d'information sur une commande grands comptes. Transfert vers le service concerné. & Un représentant du Ministère des Affaires étrangères organise un séminaire à Paris et contacte la RATP pour obtenir des cartes de transport pour les participants. & 0 & 70.14 & 70.14 \\
\midrule
135 & Demande d'itinéraire gare de Drancy gare du Nord. Précision des explications suite à un précédent appel. & Un client confus sur l'interconnexion du RER B pour se rendre à Drancy est rassuré qu'il peut voyager directement depuis Bourg-la-Reine. & 4.55 & 70.29 & 65.75 \\
\midrule
29 & Demande de renseignements sur l'achat de ticket pour des groupes scolaires. Transfert vers le service concerné. & Un représentant d'une association scolaire appelle pour comprendre la procédure de transport en commun pour les étudiants. L'agent propose une solution et transfère l'appel. & 8.51 & 72.83 & 64.32 \\
\midrule
68 & appel erroné l'appelant ayant voulu joindre l'assurance maladie au 36 46 et ayant mal tapé le numéro & Un client confond le service client de la RATP avec la Sécurité Sociale, concernant un problème de carte vitale aux Antilles. & 4.65 & 66.03 & 61.38 \\
140 & aucun abonnement de la carte ImaginR par voie postale, seulement au guichet dans toutes les gares et stations & Un client a demandé un dossier Imagine+R par courrier. L'agent a expliqué que c'est impossible mais a fourni le numéro du service pour plus d'informations. & 4.17 & 65.54 & 61.38 \\
\bottomrule
\end{tabular}
\caption{\label{tab:discrepancy_decoda_4_HGR_WL} Data points where GPT-generated summaries (\texttt{4-HGR$\rightarrow$WL}) show the greatest discrepancy in the DECODA corpus, with low ROUGE-L but high BERTScore values.}
\end{table*}

\section{Datasets: License or Terms of Use}

The DialogSum corpus used in this study comprises resources that are freely available online without copyright restrictions for academic use. 
For the DECODA dataset, the training and validation sets can be downloaded from their website\footnote{\url{https://pageperso.lis-lab.fr/benoit.favre/cccs/}} upon acceptance of the corresponding usage and sharing terms, while the test set is available only upon request from the authors.

\end{document}